\documentclass[11pt,twoside,fleqn]{article}

\usepackage[letterpaper,margin=0.68in]{geometry}
\usepackage[T1]{fontenc}
\usepackage{microtype}
\usepackage{amsmath,amssymb,amsthm,bm}
\usepackage{graphicx}
\usepackage{tikz}
\usetikzlibrary{positioning,arrows.meta}
\usepackage{authblk}
\usepackage{xcolor}
\usepackage[numbers,sort&compress]{natbib}
\providecommand{\natexlab}[1]{}\renewcommand{\natexlab}[1]{}
\usepackage[colorlinks=true,linkcolor=blue,citecolor=blue,urlcolor=blue]{hyperref}

\renewcommand{\d}{\mathrm{d}}

\newif\ifrevmarkup
\revmarkupfalse
\ifrevmarkup
  \definecolor{darkred}{RGB}{139,0,0}
  \colorlet{revlatest}{blue}
\else
  \definecolor{darkred}{RGB}{0,0,0}
  \colorlet{revlatest}{black}
  \hypersetup{hidelinks}
\fi
\newcommand{\rev}[1]{{\color{darkred}#1}}
\newenvironment{revtext}{\color{darkred}}{}
\newcommand{\revb}[1]{{\color{darkred}#1}}
\newenvironment{revbtext}{\color{darkred}}{}
\newcommand{\revc}[1]{{\color{revlatest}#1}}
\newenvironment{revctext}{\color{revlatest}}{}
\colorlet{revfix}{blue}

\title{\textbf{Score-Based Modeling of Effective Langevin Dynamics \rev{with State-Dependent Mobility}}}
\author[1]{Ludovico T. Giorgini\thanks{ludogio@mit.edu; \url{https://ludogiorgi.github.io/}}}
\affil[1]{Department of Mathematics, Massachusetts Institute of Technology, Cambridge, MA 02139, USA}
\date{}

\begin{document}

\makeatletter
\twocolumn[
\begin{@twocolumnfalse}
\maketitle

\begin{abstract}
\begin{revctext}
In a previous work {[}L.~T.~Giorgini, Phys.\ Rev.\ E \textbf{114}, L012102
(2026){]}, we introduced stochastic reduced-order models that reproduce by
construction the stationary density and short-time correlations of the observed
variables, using a constant mobility matrix and the score, the gradient of the
logarithm of this density. Here we generalize the framework to state-dependent
mobilities, calibrated so that the model also reproduces the finite-lag
correlation functions of chosen target observables. The calibration uses exact
finite-lag forms of the generalized fluctuation--dissipation theorem, which
express time derivatives of lagged correlations as averages over observed pairs
of states and are linear in the mobility. When the mobility is expanded on a
basis of known fields, its coefficients follow from a least-squares problem
involving only the stationary score, and the observables and basis are selected
by data-driven diagnostics. Otherwise, a neural-network mobility is calibrated
through the conditional score of the finite-time transition density. We validate
the approach analytically on a Cox--Ingersoll--Ross process and numerically on a
two-dimensional nonequilibrium diffusion with multiplicative noise and a
\(36\)-dimensional soft-spin Landau--Lifshitz chain. Both methods reproduce the stationary statistics and the target correlations accurately. Because the identities can be integrated between consecutive lags, so
that no time derivatives need to be estimated, the recovered drift and diffusion
are much less sensitive to coarse sampling than Kramers--Moyal estimates, which
require short sampling intervals. Calibration requires no simulation of the model and
applies to high-dimensional systems.
\end{revctext}
\end{abstract}

\vspace{1em}
\end{@twocolumnfalse}
]
\makeatother

\section{Introduction}

\begin{revtext}
Stochastic reduced-order models (ROMs) are ubiquitous in the study of
multiscale systems. By retaining a small set of resolved variables and
replacing the fast, unresolved scales with noise, they provide simplified
descriptions of systems whose full dynamics are too complex or too costly to
simulate
\cite{Zwanzig1961,Mori1965,ChorinHaldKupferman2000,MTV2,pavliotis2008multiscale,WoutersLucarini2012,WoutersLucarini2013},
and they are widely used as stochastic parameterizations, in particular in
climate science
\cite{Hasselmann1976,Penland1989,Wilks2005,KONDRASHOV201533,Berner2017,ghil2020physics,LucariniChekroun2023}.
Because this replacement is an approximation, the purpose of a ROM is not to
reproduce observed trajectories but to capture the statistical and dynamical
features of interest, typically the stationary distribution of the resolved
variables, their time-correlation functions, and their response to
perturbations
\cite{Ruelle2009,falasca2025probing,falasca2026causalrom,lucarini2026koopmanism}.

Several routes have been explored to calibrate ROMs against such targets. One
route is simulation-based calibration, in which candidate reduced models are
repeatedly integrated and their simulated statistical and dynamical features are
matched to observed targets
\cite{GourierouxMonfortRenault1993,GutmannCorander2016}.
This strategy is general, but it becomes expensive when each model evaluation
requires a long stochastic integration. A second route is local
trajectory-based reconstruction, including Kramers--Moyal and Fokker--Planck
methods, which infer drift and diffusion from short-time conditional increments
or finite-difference tendencies
\cite{MalpicaMorales2025,Siegert1998,RagwitzKantz2001,Friedrich2011,giorgini2022non,keyes2023stochastic}.
These methods are powerful when sufficiently resolved trajectories are
available, but finite-difference estimates of local tendencies and short-time
conditional moments become unreliable when observations are noisy, unevenly
sampled, or saved too coarsely relative to the intrinsic time scales
\cite{RagwitzKantz2001,Friedrich2011}. Estimators based on integrated generator
identities \cite{Krumscheid2013,Kalliadasis2015}, on lagged covariances
\cite{PenlandSardeshmukh1995}, or on eigenpairs of the
transition operator
\cite{KesslerSorensen1999,CrommelinVandenEijnden2006,CrommelinVandenEijnden2011}
do not require short sampling intervals.  However, they rely on parametric forms of the drift
and diffusion or on basis representations of the generator, which become
impractical in high dimension. A third route approximates coarse generators,
transfer operators, or lagged correlation structures through partitions,
clustering, delay-coordinate embeddings, or operator-theoretic constructions
\cite{Takens1981,GivonKupfermanStuart2004,BotvinickGreenhouse2023,BotvinickGreenhouse2025,SantosGutierrezLucariniChekrounGhil2021,KlusEtAl2020gEDMD,falasca2024data,falasca2025FDT,giorgini2025learning,giorgini2024reduced,souza2024representing_a,souza2024representing_b,souza2024modified}.
These approaches are conceptually close to the present work, but direct
reconstruction of local coefficients from state-space partitions suffers from
the curse of dimensionality. These limitations motivate an inverse formulation
in which an effective stochastic generator is constrained directly from
stationary samples and lagged pairs, without repeated forward integration during
calibration, without estimating instantaneous trajectory velocities, and without
reconstructing local coefficients by state-space clustering.

Building on earlier work
\cite{YinAo2006,MaChenFox2015,DuncanLelievrePavliotis2016}, in
Ref.~\cite{giorgini2026scorelangevin} we proposed a different strategy, in
which the target statistics are built into the model rather than verified after
calibration; this approach has since been applied to multiscale chaotic
dynamics and to cyclostationary time series
\cite{del2025integrating,giorgini2025cyclostationary}.
The construction rests on the score of the stationary density of
the resolved variables, that is, the gradient of its logarithm.
\revc{Score matching estimates it directly from samples, without estimating the
density \cite{Hyvarinen2005,Vincent2011,SongSohlDicksteinKingma2021} and has been used to
learn the scores of high-dimensional dynamical systems
\cite{giorgini_response_theory,giorgini2026kgmm}. Here we use implicit score
matching \cite{Hyvarinen2005} with networks built from local energies, whose
cost grows linearly with the dimension.}
The score is then combined with a mobility matrix that encodes the
dynamical couplings among the resolved variables: its symmetric part governs dissipation
and diffusion, while its antisymmetric part governs the probability currents
that characterize nonequilibrium systems. By construction, the resulting model
preserves the observed invariant density regardless of the mobility, which
remains free to be determined from the dynamical properties of the data. Such
models are inexpensive to calibrate, also in high dimension, because calibration
only requires a score estimate and correlation functions measured from the data,
with no integration of candidate models and no iterative adjustment against
simulated statistics.

In that work, however, we approximated the mobility with a constant matrix and
constrained it using only the short-time correlations of the state variables.
In the present work we remove this restriction.
To calibrate the
state-dependent mobility, we use the generalized
fluctuation--dissipation theorem (GFDT)
\cite{Kubo1966,MarconiPuglisiRondoniVulpiani2008,majda2009,GRITSUN,cooper2011climate,LucariniChekroun2023},
which relates the response of
a system to a perturbation to correlation functions of the unperturbed system
and, combined with score matching, has been used for response prediction and
parameter calibration
\cite{giorgini_response_theory,giorgini2025probdist,giorgini2025calibration}.
Here we use it to relate the linear response to a perturbation built from the
mobility to the time derivative of lagged correlation functions between selected
observables. The result is an identity that holds for arbitrary observables and
lags, is linear in the unknown mobility, and recovers the mean mobility of
Ref.~\cite{giorgini2026scorelangevin} for coordinate observables at vanishing
lag. It can also be integrated between consecutive lags, so that it involves
differences of observed correlations instead of their time derivatives. Unlike
Kramers--Moyal estimates, the integrated identity therefore does not require a
short sampling interval.

We illustrate the method on three systems of increasing complexity. The
first is the Cox--Ingersoll--Ross diffusion, a one-dimensional process for
which every quantity entering the calibration is known in closed form. The
second is a two-dimensional nonequilibrium diffusion with multiplicative noise,
for which the scores and the correlation functions are estimated from simulated
trajectories. The third is a periodic chain of twelve three-dimensional soft spins evolving under
stochastic Landau--Lifshitz dynamics, a \(36\)-dimensional system with relaxation times separated by two orders of magnitude.
For the latter two systems, we compare the calibrated models with
Kramers--Moyal estimates as the sampling interval of the data increases,
showing that the calibration based on the integrated identity is much
less sensitive to coarse sampling.
The paper is
organized as follows. Section~\ref{sec:method} derives the identities, two
calibration methods based on them, and the identifiability results, and summarizes the
procedure. Section~\ref{sec:results} presents the three examples.
Section~\ref{sec:conclusions} discusses implications and limitations, and the
Appendices collect derivations and implementation details.
\end{revtext}

\section{Method}
\label{sec:method}

\subsection{Problem setting and modeling objective}
\label{subsec:problem_setting_modeling_objective}

\begin{revtext}
We consider a resolved process $\bm{x}(t)\in\mathbb{R}^D$ obtained from
observations, for which we want to construct an effective stochastic model directly from
data. {\color{darkred}After removing any slow trends, we assume that the resolved
process is stationary, with density \(p_{\mathrm{ss}}(\bm{x})\).
For cyclostationary observations, we augment the resolved state with the phase of
the periodic forcing, yielding an autonomous system. In this case, the
stationarity assumption applies to the augmented process, and
\(p_{\mathrm{ss}}\) denotes its joint stationary density. We further assume
ergodicity and sufficiently rapid decay of the relevant fluctuation
correlations, so that expectations can be estimated from a sufficiently long
realization.}
\end{revtext}

We model the resolved dynamics by an It\^o diffusion of the form
\begin{equation}
    \d \bm{x}(t)
    =
    \bm{F}(\bm{x}(t))\,\d t
    +
    \sqrt{2}\,\bm{\Sigma}(\bm{x}(t))\,\d \bm{W}_t,
    \label{eq:method_sde_full}
\end{equation}
where $\bm{F}:\mathbb{R}^D\to\mathbb{R}^D$ is the drift,
$\bm{\Sigma}:\mathbb{R}^D\to\mathbb{R}^{D\times D}$ is the noise-amplitude matrix, and $\bm{W}_t$ is a standard $D$-dimensional Wiener process. We denote the diffusion tensor by
\begin{equation}
    \bm{D}(\bm{x})
    :=
    \bm{\Sigma}(\bm{x})\bm{\Sigma}(\bm{x})^T .
    \label{eq:diffusion_tensor_definition}
\end{equation}
\begin{revtext}
Both \(\bm{F}\) and \(\bm{D}\) are unknown.
We choose them so that the model reproduces two classes of statistics of the
observed process. The first is the stationary density \(p_{\mathrm{ss}}\), which
determines all single-time statistics of the resolved variables. The second is a
set of target dynamical features, which we take to be lagged correlation
functions
\(\bm{C}_{f,g,\mathrm{obs}}(t)=\langle f(\bm{x}(t))\,g(\bm{x}(0))^T\rangle_{\mathrm{obs}}\)
of target observables \(f\) and \(g\), for a prescribed range of lags
\(t\). Here \(\langle\cdot\rangle_{\mathrm{obs}}\) denotes an expectation
over the observed lagged pairs, estimated in practice by an empirical average
over them. The target observables are dictated by the
application and may include state variables and nonlinear
functions of them, such as an energy or a squared amplitude.
Unsubscripted angle brackets denote the corresponding expectations over
the stationary process, and \(\langle\cdot\rangle_{p_{\mathrm{ss}}}\) denotes the
average \(\int(\cdot)\,p_{\mathrm{ss}}\,\d\bm{x}\) of a function of a single
state. The modeling objective is thus to choose \(\bm{F}\) and \(\bm{D}\)
such that the stationary density and the target correlations of the model match
the observed ones,
\begin{equation}
\begin{aligned}
    p_{\mathrm{model}}(\bm{x})
    &\approx
    p_{\mathrm{ss}}(\bm{x}),
    \\
    \bm{C}_{f,g,\mathrm{model}}(t)
    &\approx
    \bm{C}_{f,g,\mathrm{obs}}(t).
\end{aligned}
    \label{eq:modeling_objective}
\end{equation}
\end{revtext}

To reproduce the stationary density, we use its stationary score,
\begin{equation}
    \bm{s}(\bm{x})
    :=
    \nabla \log p_{\mathrm{ss}}(\bm{x}).
    \label{eq:score_definition_method}
\end{equation}
\begin{revtext}
Following
Ref.~\cite{giorgini2026scorelangevin}, a stationary diffusion with invariant
density \(p_{\mathrm{ss}}\) can be written in the score-based form
\end{revtext}
\begin{equation}
\begin{aligned}
    \d \bm{x}(t)
    &=
    \left[
        \bm{M}(\bm{x}(t))\bm{s}(\bm{x}(t))
        +
        \nabla\cdot\bm{M}(\bm{x}(t))
    \right]\d t
    \\
    &\quad
    +
    \sqrt{2}\,\bm{\Sigma}(\bm{x}(t))\,\d \bm{W}_t,
\end{aligned}
    \label{eq:compact_sde_full}
\end{equation}
where \rev{the mobility is decomposed as}
\begin{equation}
\begin{aligned}
    \bm{M}(\bm{x})
    &=
    \bm{D}(\bm{x})
    +
    \bm{R}(\bm{x}),
    \\
    \bm{R}(\bm{x})^T
    &=
    -\bm{R}(\bm{x}),
    \\
    \bm{D}(\bm{x})
    &=
    \frac{\bm{M}(\bm{x})+\bm{M}(\bm{x})^T}{2}
    \succeq 0 .
\end{aligned}
    \label{eq:mobility_decomposition_method}
\end{equation}
Here $(\nabla\cdot\bm{M})_i=\sum_{j=1}^D\partial_{x_j}M_{ij}$. 
\begin{revtext}
{\color{darkred}The symmetric part \(\bm{D}\) determines the diffusion and the
reversible part of the drift, while the antisymmetric part \(\bm{R}\) can sustain
stationary probability currents and thus allows the model to describe
nonequilibrium dynamics.}
Eq. \eqref{eq:compact_sde_full} imposes the steady-state density by construction for every admissible state-dependent mobility, that is, one that is
sufficiently regular, has a positive semidefinite symmetric part, and
generates no probability flux through the boundary of the state space~\cite{MaChenFox2015,giorgini2026scorelangevin}. The stationary score is learned
once from samples
(Appendix~\ref{app:score_estimation}), and the remaining
unknown is the mobility that determines both the drift and the
diffusion.
\end{revtext}

\subsection{\texorpdfstring{\rev{Calibration of the mobility}}{Calibration of the mobility}}
\label{subsec:mobility_from_correlation_constraints}
\label{subsec:gfdt_interpretation_correlation_constraint}

\begin{revtext}
The remaining objective is to calibrate the mobility \(\bm M(\bm x)\) so that the model reproduces the target correlations in Eq.~\eqref{eq:modeling_objective}. We derive a weak identity that expresses their time derivatives as averages over stationary lagged pairs. This identity leads to two calibration procedures: linear regression when the mobility is expanded on known matrix fields, and nonlinear optimization when it is represented by a neural network. We use scalar observables below; vector observables are treated componentwise.

Writing \(p=p_{\mathrm{ss}}\), the backward generator of Eq.~\eqref{eq:compact_sde_full} and its adjoint in \(L^2(p)\) are
\begin{align}
\mathcal L_{\bm M}u
=
\frac{1}{p}\nabla\cdot\!\left(p\bm M^T\nabla u\right)\label{eq:generator_divergence_form}\\
\mathcal L_{\bm M}^{\dagger}v
=
\frac{1}{p}\nabla\cdot\!\left(p\bm M\nabla v\right).
\label{eq:adjoint_generator}
\end{align}
For admissible smooth observables, integration by parts gives
\begin{equation}
\begin{aligned}
\left\langle(\mathcal L_{\bm M}u)v\right\rangle_p
&=\left\langle u\,\mathcal L_{\bm M}^{\dagger}v\right\rangle_p
\\
&=\Bigl\langle u\Bigl[(\bm M^T\bm s+\nabla\cdot\bm M^T)\cdot\nabla v
\\
&\qquad\qquad+\bm D:\nabla^2v\Bigr]\Bigr\rangle_p.
\end{aligned}
\label{eq:weak_generator_identity}
\end{equation}
Let \(K_t^{\bm M}f(\bm x_0)=
\mathbb E_{\bm M}[f(\bm x_t)\mid\bm x_0]\) denote the Koopman semigroup of the model. Applying this identity to \(u=K_t^{\bm M}f\) and \(v=g\) yields
\begin{equation}
\dot C_{f,g}(t)
=
\left\langle
f(\bm x_t)\,
(\mathcal L_{\bm M}^{\dagger}g)(\bm x_0)
\right\rangle.
\label{eq:Cdot_as_gfdt_response}
\end{equation}
The correlation derivative and the lagged average are evaluated from observations, turning this relation into a linear constraint on \(\bm M\).

The identity can be interpreted in terms of linear response theory. By the generalized fluctuation--dissipation theorem~\cite{Agarwal1972,FalcioniIsolaVulpiani1990,MarconiPuglisiRondoniVulpiani2008}, the response of \(f\) to a drift impulse \(\varepsilon\bm b(\bm x)\delta(t)\) is determined by its correlation with \(B_{\bm b}=-p^{-1}\nabla\cdot(p\bm b)\). For \(\bm b=-\bm M\nabla g\), this conjugate observable is \(\mathcal L_{\bm M}^{\dagger}g\). The correlation derivative in Eq.~\eqref{eq:Cdot_as_gfdt_response} therefore equals the response to a mobility-weighted impulse. At equilibrium, with \(\bm R=0\), \(\bm D=\beta^{-1}\bm I\), and \(p\propto e^{-\beta U}\), the response to a field coupled through \(U\mapsto U-h(t)g\) reduces to the classical relation~\cite{CallenWelton1951,Kubo1957,Kubo1966}
\begin{equation}
\mathcal R_{f,g}^{\mathrm{eq}}(t)
=
-\beta\,\frac{\mathrm d}{\mathrm dt}C_{f,g}^{\mathrm{eq}}(t),
\qquad t>0.
\label{eq:method_classical_equilibrium_fdt}
\end{equation}

Imposing the weak identity only for the target pair \(f,g\) does not necessarily reproduce the target correlations after integrating the model. The chosen observables may not span a space preserved by time evolution and then the conditional mean of a target observable at a later time can lie outside their span. The weak identities have not been imposed for these additional functions, so matching the target identities alone need not determine the correlations generated by the model and additional observables may therefore be required to constrain the generator. We introduce two finite libraries,
\begin{equation}
\begin{aligned}
\mathcal U&=\operatorname{span}\{\psi_m\}_{m=1}^{N_\psi},
\\
\mathcal V&=\operatorname{span}\{\phi_n\}_{n=1}^{N_\phi},
\\
f&\in\mathcal U,\quad g\in\mathcal V,
\end{aligned}
\label{eq:observable_family_method}
\end{equation}
evaluated at the future and initial times, respectively. Their observed correlations are
\begin{equation}
C_{mn,\mathrm{obs}}(t)
=
\left\langle\psi_m(\bm x_t)\phi_n(\bm x_0)\right\rangle_{\mathrm{obs}},
\label{eq:Cm_definition}
\end{equation}
and the calibration imposes
\begin{equation}
\dot C_{mn,\mathrm{obs}}(t)
\approx
\left\langle
\psi_m(\bm x_t)
(\mathcal L_{\bm M}^{\dagger}\phi_n)(\bm x_0)
\right\rangle_{\mathrm{obs}}
\label{eq:stationary_score_identity}
\end{equation}
for selected pairs and positive lags.

When the sampling interval resolves the correlation functions, their time derivatives are estimated by differentiating local-polynomial fits to the empirical correlations. Otherwise, Eq.~\eqref{eq:stationary_score_identity} is integrated between consecutive lags, so that differences of observed correlations replace their derivatives (Appendix~\ref{app:fixed_basis_details}). The mean mobility can then be fitted together with the rest of the mobility, which avoids estimating the initial slope of the coordinate correlations, itself a derivative at zero lag.
\end{revtext}

\subsubsection{\texorpdfstring{\rev{Calibration on a fixed basis}}{Calibration on a fixed basis}}
\label{subsec:fixed_basis_calibration}

\begin{revtext}
Suppose that the mobility belongs to, or can be approximated by, a finite family of smooth matrix fields:
\begin{equation}
\bm M_{\bm c}(\bm x)
=
\sum_{k=1}^{K}c_k\bm H_k(\bm x).
\label{eq:basis_ansatz_uncentered}
\csname ltx@label\endcsname{eq:basis_ansatz}
\end{equation}
The coefficients must be chosen so that the symmetric part of the resulting mobility is positive semidefinite.

For each observable \(\phi_n\), define the conjugate observables
\begin{equation}
q_{kn}(\bm x)
=
(\mathcal L_{\bm H_k}^{\dagger}\phi_n)(\bm x),
\label{eq:basis_response_functions}
\end{equation}
computed from the stationary score and the analytic derivatives of the basis fields and observables. Linearity in the mobility then gives the system \(\bm B\bm c\approx\bm y\), with
\begin{equation}
\begin{aligned}
B_{\alpha k}&=\left\langle\psi_m(\bm x_{t_\ell})q_{kn}(\bm x_0)\right\rangle_{\mathrm{obs}},\\
y_\alpha&=\dot C_{mn,\mathrm{obs}}(t_\ell),\qquad\alpha=(\ell,m,n).
\end{aligned}
\label{eq:library_linear_system}
\end{equation}
All entries are averages over observed data. The coefficients are obtained by weighted least squares,
\begin{equation}
\widehat{\bm c}
=
\operatorname*{arg\,min}_{\bm c}
\left\|
\bm W^{1/2}
(\widehat{\bm B}\bm c-\widehat{\bm y})
\right\|_2^2,
\label{eq:basis_wls}
\end{equation}
with the constraint that the symmetric part of the mobility has no negative eigenvalues. In practice, we impose this constraint on a grid of states spanning the region sampled by the data and solve the resulting convex constrained least-squares problem, as detailed in Appendix~\ref{app:fixed_basis_details}. Hats denote finite-sample estimates. The weight matrix \(\bm W\) is diagonal, with entries \(W_{\alpha\alpha}=1/\sigma_\alpha^2\), where \(\sigma_\alpha\) is the jackknife estimate of the standard error of \(\widehat y_\alpha\)~\cite{Efron1982}.

The observables \(\phi_n\), evaluated at the initial state \(\bm x_0\), determine which combinations of the coefficients the calibration can distinguish. A combination \(\sum_k a_k\bm H_k\) for which \(\sum_k a_kq_{kn}=0\) for every \(n\) cannot be identified from the calibration: adding it to the mobility leaves the right-hand sides of all calibration equations unchanged, for any choice of future observables and lags. Such combinations are excluded when the Gram matrix
\begin{equation}
[\bm G_\phi]_{k\ell}
=
\sum_n\left\langle q_{kn}q_{\ell n}\right\rangle_p
\label{eq:phi_gram}
\end{equation}
is positive definite. Any coefficient combination that changes neither the drift nor the diffusion represents a gauge freedom: changing its coefficient changes the mobility representation but leaves the stochastic dynamics unchanged. Such combinations cannot be identified from observations and must be removed before fitting.

Equation~\eqref{eq:weak_generator_identity} shows that the adjoint acts on
\(\phi_n\) only through its first and second derivatives. For the coordinate
functions \(\phi_i=x_i\), its action is therefore
\begin{equation}
    \mathcal L_{\bm H_k}^{\dagger}x_i
    =(\bm H_k^T\bm s+\nabla\cdot\bm H_k^T)_i
    =:(\bm a_k)_i.
\label{eq:basis_coordinate_response}
\end{equation}
Coordinates therefore probe $\bm H_k$ only through the drift $\bm a_k$ it induces, so basis elements with the same drift are indistinguishable.
If the resulting $\bm G_\phi$ is singular, we add the quadratic functions $x_ix_j$, whose response also contains $\operatorname{sym}(\bm H_k)$ directly:
\begin{equation}
\begin{aligned}
    \mathcal L_{\bm H_k}^{\dagger}(x_ix_j)
    &=x_j(\bm a_k)_i+x_i(\bm a_k)_j
    \\
    &\quad+2\operatorname{sym}(\bm H_k)_{ij}.
\end{aligned}
\label{eq:basis_quadratic_response}
\end{equation}
Here \(\operatorname{sym}\bm H_k=(\bm H_k+\bm H_k^T)/2\). The additional
term distinguishes changes in the diffusion. The first-derivative and diffusion
terms completely specify the operator; coordinates and quadratics together
therefore detect every basis combination that changes the generator.

The future observables \(\psi_m\) and the lags must then provide enough
independent equations to determine the coefficients, which requires the columns
of \(\bm B\) to be linearly independent. We include the conjugate observables
\(q_{kn}\) themselves among the future observables. The resulting entries of
\(\bm B\) include the correlations
\(\langle q_{jn}(\bm x_t)q_{kn}(\bm x_0)\rangle\). Summing these entries over
\(n\) gives a square matrix,
\begin{equation}
    S_{jk}(t)=\sum_n\left\langle
       q_{jn}(\bm x_t)q_{kn}(\bm x_0)\right\rangle.
\label{eq:basis_response_lag_matrix}
\end{equation}
Each row of \(\bm S(t)\) is a sum of rows of \(\bm B\). Since
\(\bm S(0)=\bm G_\phi\) and \(\bm S(t)\to\bm G_\phi\) as \(t\to0^+\),
positive definiteness of \(\bm G_\phi\) ensures that \(\bm B\) has linearly
independent columns at sufficiently short positive lags. At longer lags,
correlations decay, reducing the information available to distinguish the
coefficients. Finite-sample errors can also make linearly dependent columns
appear independent, suggesting that coefficients are identifiable when they
are not. We therefore check whether the positive definiteness of
\(\bm G_\phi\) and the column independence of \(\bm B\) remain supported
after accounting for sampling uncertainty. This uncertainty is estimated by
repeating the calculations with one trajectory omitted at a time, as described
in Appendix~\ref{app:fixed_basis_details}.

For a fixed mobility basis, the libraries are therefore constructed from the
basis fields themselves. We start from a candidate family \(\{\bm H_k\}\),
take the coordinates as \(\phi_n\), and check \(\bm G_\phi\). If it is not
positive definite, we add the quadratic observables. With exact scores and
averages, there are combinations of basis functions \(\bm H_{\bm a}=\sum_k a_k\bm H_k\) that leave the dynamics unchanged. They satisfy
\begin{equation}
\begin{aligned}
    \bm H_{\bm a}(\bm x)&=\frac{\bm Q(\bm x)}{p(\bm x)},
    \\
    \bm Q^T&=-\bm Q,\qquad\nabla\cdot\bm Q=0.
\end{aligned}
\label{eq:basis_mobility_gauge}
\end{equation}
Its symmetric part and its drift contribution
\(\bm H_{\bm a}\bm s+\nabla\cdot\bm H_{\bm a}\) vanish, so it changes neither
the drift nor the diffusion. Such combinations must be removed because their
coefficients cannot be identified. Small estimated eigenvalues must first be checked
against sampling uncertainty before identifying an exact null direction.

We start from a candidate mobility basis, calibrate its coefficients, and
integrate the resulting model to compare its target correlations with the
observations. If agreement is insufficient, we enlarge the basis and repeat
the calibration. We then progressively remove terms to identify a minimal
basis that preserves the required accuracy. After selecting the reduced basis,
we refit its coefficients and integrate the model to verify that it reproduces
the target correlations to the required accuracy. Enlargement and reduction
are repeated as needed. Details of the basis-selection procedure are given in
Appendix~\ref{app:fixed_basis_details}.
Table~\ref{tab:fixed_basis_algorithm} summarizes the procedure.
\end{revtext}

\begin{table*}[t]
\begin{revtext}
\centering
\caption{Calibration on a fixed mobility basis.}
\label{tab:fixed_basis_algorithm}
\small
\renewcommand{\arraystretch}{1.18}
\begin{tabular}{@{}p{0.05\textwidth}p{0.91\textwidth}@{}}
\hline
\textbf{Step} & \textbf{Operation} \\
\hline
1 & Choose an initial family \(\{\bm H_k\}\) for
\(\bm M_{\bm c}=\sum_kc_k\bm H_k\). \\
2 & Start with \(\phi_i=x_i\), compute
\(q_{kn}=\mathcal L_{\bm H_k}^{\dagger}\phi_n\), and evaluate
\(\bm G_\phi\). \\
3 & If \(\bm G_\phi\) is not positive definite, add \(x_ix_j\).
Remove any remaining combinations of the form
\eqref{eq:basis_mobility_gauge}, which leave the dynamics unchanged.
Distinguish exact null directions from sampling uncertainty. \\
4 & Take \(\psi_m=q_{kn}\), retain the target observables \(f\) and \(g\)
in their respective libraries, and choose informative positive lags. \\
5 & Estimate \(\bm y\) and \(\bm B\), and check that the columns of
\(\bm B\) are independent relative to their sampling uncertainty. \\
6 & Solve the weighted least-squares problem~\eqref{eq:basis_wls},
subject to admissibility. \\
7 & Integrate the calibrated model and compare its target correlations
with the observations. If agreement is insufficient, enlarge
\(\{\bm H_k\}\) and return to step 2. \\
8 & Progressively remove terms to identify a minimal basis that preserves
the required accuracy (Appendix~\ref{app:fixed_basis_details}). Refit the
selected reduced basis and integrate the model to verify its target
correlations. Repeat enlargement and reduction as needed. \\
\hline
\end{tabular}
\end{revtext}
\end{table*}

\subsubsection{\texorpdfstring{\rev{Neural mobility calibration}}{Neural mobility calibration}}
\label{subsec:neural_conditional_score}
\label{subsec:mean_mobility_and_corrections}
\label{subsec:model_identifiability}

\begin{revtext}
When no suitable finite family is known, we represent the mobility by a neural network with parameters \(\theta\). These parameters enter nonlinearly. We ensure that the symmetric part of the mobility has no negative eigenvalues at any state by writing
\begin{equation}
\begin{aligned}
\bm M_\theta(\bm x)
&=
\bm L_\theta(\bm x)\bm L_\theta(\bm x)^T
+\varepsilon\bm I
+\bm R_\theta(\bm x),
\\
\bm R_\theta^T&=-\bm R_\theta,
\end{aligned}
\label{eq:neural_M_DR_parametrization}
\end{equation}
where \(\bm L_\theta\) is lower triangular and the constant \(\varepsilon>0\) bounds the eigenvalues of the diffusion tensor \(\bm D_\theta=\bm L_\theta\bm L_\theta^T+\varepsilon\bm I\) from below, so that the noise remains nondegenerate where \(\bm L_\theta\bm L_\theta^T\) is singular. The architecture is chosen to encode the locality and symmetries of the modeled system, for example by letting each block of the mobility depend only on the variables that interact directly with the corresponding components, and by making the network equivariant under the transformations that leave the dynamics invariant.

In the fixed-basis calibration, the prescribed matrix fields provide a systematic construction of the observable libraries. For a neural mobility, we must choose the analytic functions \(\phi_n\) and \(\psi_m\) used to constrain the network. A sufficient condition under which enforcing the weak identities ensures that the calibrated model reproduces the target correlations is that the observed conditional evolution forms a linear semigroup on \(\mathcal U\) and that \(\mathcal V\) contains the generator residuals defined in Appendix~\ref{app:finite_library_exact_recovery}.

The semigroup condition provides a useful criterion for choosing the future observables: starting from the targets, the library can be enriched so that a common linear evolution within its span approximates the observed conditional means over the fitted lags. Exact closure may nevertheless be difficult to achieve in high dimensions. The condition on the initial observables is harder to implement, since the generator residuals depend on the neural mobility and its spatial derivatives and change during optimization. In practice, we use fixed analytic libraries containing the target observables, supplemented where needed by additional functions. We favor functions whose finite-lag correlation derivatives are large relative to their sampling uncertainty and avoid nearly duplicate correlation curves. We assess the libraries by integrating the calibrated model and comparing its target correlations with the observations.

Direct calibration with Eq.~\eqref{eq:stationary_score_identity} would require the spatial divergence of the mobility network and the first and second derivatives of \(\phi_n\), as shown in Eq.~\eqref{eq:weak_generator_identity}. Training then involves mixed derivatives with respect to the state and network parameters, increasing computational and memory costs. Moreover, the second derivatives of \(\phi_n\) are difficult to evaluate when the observables are not smooth. We therefore derive a conditional-score identity that removes the spatial derivatives of the mobility and requires only first derivatives of \(\phi_n\). We first introduce the mean mobility,
\begin{equation}
\bm\Phi
:=
\left\langle\bm M\right\rangle_p
=
-\dot{\bm C}_{\bm x,\bm x}(0^+),
\label{eq:Phi_from_Cdot_zero_method}
\csname ltx@label\endcsname{eq:mobility_mean_fluctuation_decomposition}
\end{equation}
where the last equality follows from Eq.~\eqref{eq:Cdot_as_gfdt_response} at \(t=0^+\) with \(f=x_a\) and \(g=x_b\), after one integration by parts. The estimate \(\widehat{\bm\Phi}=-\widehat{\dot{\bm C}}_{\bm x,\bm x}(0^+)\) is obtained by extrapolating to \(t=0^+\) the slope of the coordinate correlations at the first positive lags (Appendix~\ref{app:local_polynomial_cdot_estimation}), and it recovers the constant closure of Ref.~\cite{giorgini2026scorelangevin}. For the neural model, we write
\(\delta\bm M_\theta=\bm M_\theta-\widehat{\bm\Phi}\) and calibrate the correction.

Let \(p_t^{\mathrm{obs}}(\bm x_t\mid\bm x_0)\) denote the observed conditional density and define its score with respect to the initial state,
\begin{equation}
\bm s_{t|0}(\bm x_t\mid\bm x_0)
=
\nabla_{\bm x_0}
\log p_t^{\mathrm{obs}}(\bm x_t\mid\bm x_0).
\label{eq:conditional_score_method_section}
\end{equation}
Using \(\nabla_{\bm x_0}p_t^{\mathrm{obs}}=p_t^{\mathrm{obs}}\bm s_{t|0}\), we can express the gradient of the conditional mean with respect to the initial state as
\begin{equation}
\nabla K_t^{\mathrm{obs}}\psi_m(\bm x_0)
=
\mathbb E_{\mathrm{obs}}\!\left[
\psi_m(\bm x_t)\bm s_{t|0}(\bm x_t\mid\bm x_0)
\,\middle|\,\bm x_0
\right].
\label{eq:grad_Kt_conditional_score_method}
\end{equation}
This equation gives the gradient of the evolved observables directly from lagged pairs, without approximating the Koopman semigroup by eigenfunctions or other finite-dimensional representations and differentiating the approximation~\cite{Koopman1931,Mezic2005,WilliamsKevrekidisRowley2015,KlusNuskeKoltaiWuKevrekidisSchutteNoe2018,SantosGutierrezLucariniChekrounGhil2021,ZagliColbrookLucariniMezicMoroney2026,lucarini2026koopmanism}. The conditional score is learned once from lagged pairs by \revc{implicit score matching} (Appendix~\ref{app:conditional_score}) and reused for different observable libraries at the same lags.

An integration by parts of equation~\eqref{eq:grad_Kt_conditional_score_method} gives, for any smooth mobility \(\bm M\),
\begin{equation}
\begin{aligned}
\mathcal A_{mn,t}[\bm M]
&:=
\left\langle
\psi_m(\bm x_t)
(\mathcal L_{\bm M}^{\dagger}\phi_n)(\bm x_0)
\right\rangle_{\mathrm{obs}}
\\
&=
-\Bigl\langle
\psi_m(\bm x_t)\,
\bm s_{t|0}^{T}
\\
&\qquad\times
\bm M(\bm x_0)\,
\nabla\phi_n(\bm x_0)
\Bigr\rangle_{\mathrm{obs}}.
\end{aligned}
\label{eq:central_conditional_score_identity}
\csname ltx@label\endcsname{eq:sample_correlation_derivative_operator_method}
\end{equation}
The mobility now appears without spatial derivatives, and \(\phi_n\) enters only through its gradient.

The empirical constraints become
\begin{equation}
\widehat{\mathcal A}_{mn,t}[\delta\bm M_\theta]
\approx
\widehat y_{mn}(t),
\label{eq:neural_empirical_constraints}
\end{equation}
where
\begin{equation}
\widehat y_{mn}(t)
=
\widehat{\dot C}_{mn,\mathrm{obs}}(t)
-
\left\langle
\psi_m(\bm x_t)
(\mathcal L_{\widehat{\bm\Phi}}^{\dagger}\phi_n)(\bm x_0)
\right\rangle_{\mathrm{obs}}.
\label{eq:deltaM_residual_constraint}
\end{equation}
We determine \(\theta\) by minimizing
\begin{equation}
\begin{aligned}
\mathcal J(\theta)
&=
\sum_{t\in\mathcal T}\sum_{m,n}
w_{mn,t}
\left|
\widehat{\mathcal A}_{mn,t}[\delta\bm M_\theta]
-\widehat y_{mn}(t)
\right|^2
\\
&\quad+
\lambda_{\mathrm{mean}}
\left\|
\left\langle\bm M_\theta\right\rangle_{\mathrm{obs}}
-\widehat{\bm\Phi}
\right\|_F^2
+\lambda_{\mathrm{reg}}\mathcal P(\theta).
\end{aligned}
\label{eq:deltaM_neural_loss}
\end{equation}
The first term fits the averaged weak constraints; the remaining terms constrain the mean mobility and provide optional regularization. No spatial derivative of the mobility network and no simulation of the model enter this loss. As in the integrated identity~\eqref{eq:app_integrated_identity} of the fixed-basis calibration, integrating both sides of Eq.~\eqref{eq:neural_empirical_constraints} over \([t_{\ell-1},t_\ell]\) replaces the correlation derivative \(\widehat{\dot C}_{mn,\mathrm{obs}}\) with the difference \(\widehat C_{mn,\mathrm{obs}}(t_\ell)-\widehat C_{mn,\mathrm{obs}}(t_{\ell-1})\), and the remaining averages with their integrals over the same interval. This form is preferable when the data are sampled too coarsely for local-polynomial fits to resolve the correlation derivatives. In that case, \(\bm M_\theta\) can also be trained directly, without \(\widehat{\bm\Phi}\) and without the mean-mobility penalty.

Here \(\mathcal T\) is the set of fitted lags, and \(w_{mn,t}>0\) are prescribed weights determining the relative contribution of each observable pair and lag to the fit. The coefficient \(\lambda_{\mathrm{mean}}\geq0\) controls the penalty for deviations of the model's mean mobility from \(\widehat{\bm\Phi}\), while \(\lambda_{\mathrm{reg}}\geq0\) controls the strength of the optional regularization penalty \(\mathcal P(\theta)\). The notation \(\|\cdot\|_F\) denotes the Frobenius norm. The weights and penalties used in each example are specified in
Appendix~\ref{app:examples_details}.

Finally, we integrate the calibrated model and compare its target correlations with the observations. Correlations
excluded from calibration provide additional tests. If agreement is insufficient, we enrich the libraries and repeat calibration and simulation.

Table~\ref{tab:neural_mobility_algorithm} summarizes the procedure.
\end{revtext}

\begin{table*}[t]
\begin{revtext}
\centering
\caption{Calibration of a neural mobility with fixed analytic observable libraries.}
\label{tab:neural_mobility_algorithm}
\small
\renewcommand{\arraystretch}{1.18}
\begin{tabular}{@{}p{0.05\textwidth}p{0.91\textwidth}@{}}
\hline
\textbf{Step} & \textbf{Operation} \\
\hline
1 & Choose fixed analytic libraries \(\{\psi_m\}\) and \(\{\phi_n\}\)
containing the target observables \(f\) and \(g\), respectively. Favor
correlation derivatives that are large relative to their sampling uncertainty,
and avoid nearly duplicate correlation curves. Select positive lags. \\
2 & Estimate the stationary score, the correlation derivatives, and
\(\widehat{\bm\Phi}=-\widehat{\dot{\bm C}}_{\bm x,\bm x}(0^+)\).
The last quantity requires extrapolating the slope from positive lags.
In the integrated form, replace the correlation derivatives with the
correlation differences between consecutive lags. \\
3 & Learn the conditional score from observed lagged pairs at the selected lags. \\
4 & Form \(\widehat y_{mn}(t)\) using
Eq.~\eqref{eq:deltaM_residual_constraint} and evaluate the empirical operator
on \(\delta\bm M_\theta=\bm M_\theta-\widehat{\bm\Phi}\). In the
integrated form, integrate both quantities over \([t_{\ell-1},t_\ell]\). \\
5 & Minimize \(\mathcal J(\theta)\) in Eq.~\eqref{eq:deltaM_neural_loss},
using the parametrization~\eqref{eq:neural_M_DR_parametrization} to keep the
symmetric part of the mobility positive definite. \\
6 & Integrate the calibrated model and compare its target correlations with
the observations. Correlations excluded from calibration provide additional tests. \\
7 & If the target accuracy is not reached, enrich the libraries and recompute
the correlation derivatives or differences in step 2. Reuse the score estimates at the same
lags; if the lag set is enlarged, also learn the conditional score at the
new lags. Recompute the quantities in step 4 and repeat steps 5--6. \\
\hline
\end{tabular}
\end{revtext}
\end{table*}

\section{Results}
\label{sec:results}

\begin{revtext}
\revb{We consider three systems of increasing complexity to illustrate how the
proposed framework constructs effective Langevin dynamics from data. We begin with the one-dimensional
Cox--Ingersoll--Ross diffusion, to which we apply only the fixed-basis
calibration and for which all quantities entering it are available in closed
form. We then
consider a two-dimensional nonreversible diffusion with affine multiplicative
noise and a periodic chain of soft spins governed by stochastic Landau--Lifshitz
dynamics, to which we apply both calibrations. For these two systems, the scores and
correlation derivatives used in the calibration are estimated from simulated
trajectories, demonstrating the practical implementation of the method with
finite data.}
\end{revtext}

\subsection{Analytic warmup: the Cox--Ingersoll--Ross square-root diffusion}
\label{subsec:cir_benchmark}

\begin{revtext}
\begin{revbtext}
We begin with the Cox--Ingersoll--Ross (CIR) diffusion
\cite{CoxIngersollRoss1985,Feller1951}
\begin{equation}
    \d X_t
    =
    \kappa(\mu-X_t)\,\d t
    +
    \sqrt{2\gamma X_t}\,\d W_t,
    \label{eq:cir_sde_results}
\end{equation}
with \(\kappa,\mu,\gamma>0\). We assume \(\nu=\kappa\mu/\gamma>2\), so that the
process does not reach the origin and all the averages below are finite.

All the quantities that enter the calibration are known in closed form. We list
them here and derive them in Appendix~\ref{app:cir_benchmark_details}. The
stationary density is a Gamma density with shape \(\nu\) and rate
\(\beta=\kappa/\gamma\). Its score and moments are
\begin{equation}
    s(x)=\frac{\nu-1}{x}-\beta,
    \qquad
    \mu_a=\langle X^a\rangle_{p_{\mathrm{ss}}}=\frac{\Gamma(\nu+a)}{\Gamma(\nu)\beta^a},
    \label{eq:cir_stationary_score_results}
\end{equation}
for \(a>-\nu\), with mean \(\mu_1=\mu\) and variance \(V=\nu/\beta^2\). With this
score, Eq.~\eqref{eq:cir_sde_results} takes the score-based
form~\eqref{eq:compact_sde_full} with the mobility \(M(x)=\gamma x\), because
\(Ms+M'=\kappa(\mu-x)\) and the diffusion coefficient equals \(M\). This is the
mobility that the calibration must recover. The process is reversible, and its
conditional mean is affine, \(\mathbb{E}[X_t\mid X_0=x]=\mu+z(x-\mu)\) with
\(z=e^{-\kappa t}\). The lagged correlation of any observable \(f\) with the
coordinate and its time derivative are
\begin{equation}
\begin{aligned}
    C_{f,x}(t)&=\mu\langle f\rangle_{p_{\mathrm{ss}}}+z\langle(x-\mu)f\rangle_{p_{\mathrm{ss}}},
    \\
    \dot C_{f,x}(t)&=-\kappa z\langle(x-\mu)f\rangle_{p_{\mathrm{ss}}}.
\end{aligned}
    \label{eq:cir_correlations_results}
\end{equation}
For \(f=x\), the mean mobility of Eq.~\eqref{eq:Phi_from_Cdot_zero_method} is
\(\Phi=-\dot C_{x,x}(0^+)=\kappa V=\gamma\mu\).

The calibration framework of Sec.~\ref{subsec:fixed_basis_calibration} uses
these quantities estimated from data. As a proof of concept, we assume instead
that they are known exactly, and we show how the calibration derives the
mobility from them without assuming that it is linear in \(x\).

We expand the mobility in centered monomials up to an arbitrary degree \(K\),
\begin{equation}
    M_{\bm{c}}(x)=\Phi+\sum_{k=1}^{K}c_kH_k(x),
    \qquad
    H_k(x)=x^k-\mu_k.
    \label{eq:cir_polynomial_ansatz}
\end{equation}
The CIR mobility \(\gamma x=\Phi+\gamma H_1\) corresponds to
\(\bm{c}=\gamma\bm{e}_1\), with \(\bm{e}_1=(1,0,\ldots,0)^T\). For the
initial-time observable \(\phi=x\), the adjoint
generator~\eqref{eq:adjoint_generator} reduces to
\(\mathcal{L}^\dagger_Hx=Hs+H'\), so the
conjugate observables~\eqref{eq:basis_response_functions} are \(q_k=H_ks+H_k'\).
These functions are linearly independent (Appendix~\ref{app:cir_benchmark_details}),
so their Gram matrix \(\bm{G}_x\) of Eq.~\eqref{eq:phi_gram} is positive definite
for every finite \(K\). The coordinate therefore suffices as the initial-time
library. A future library consisting only of \(\psi=x\), however, cannot determine
the coefficients. Integration by parts gives
\(\langle q_k\rangle_{p_{\mathrm{ss}}}=\langle xq_k\rangle_{p_{\mathrm{ss}}}=0\);
together with reversibility and Eq.~\eqref{eq:cir_correlations_results}, these
identities imply \(C_{x,q_k}(t)=C_{q_k,x}(t)=0\). Thus, all entries of the
calibration matrix \(\bm{B}\) vanish for this choice. Following
Sec.~\ref{subsec:fixed_basis_calibration}, we therefore choose the nonlinear future
observables \(\psi_j=q_j\), \(j=1,\ldots,K\). Since
\(\mathcal{L}^\dagger_\Phi x=\Phi s\), the weak calibration identity becomes
\begin{equation}
    B_{jk}(t)=C_{q_j,q_k}(t),
    \qquad
    y_j(t)=\dot C_{q_j,x}(t)-\Phi\,C_{q_j,s}(t).
    \label{eq:cir_fixed_basis_system}
\end{equation}
The vanishing correlations \(C_{q_j,x}(t)=0\) and the relation between the score
and the conjugate observables give
\begin{equation}
    \bm{y}(t)=\frac{\Phi}{\mu}\,\bm{B}(t)\bm{e}_1
    \label{eq:cir_fixed_basis_target}
\end{equation}
for all \(t\), as derived in Appendix~\ref{app:cir_benchmark_details}.
Reversibility and linear independence of the conjugate observables ensure that
\(\bm{B}(t)\) is positive definite at every finite positive lag. The calibration
system therefore has the unique solution
\begin{equation*}
    \bm{c}=\frac{\Phi}{\mu}\bm{e}_1=\gamma\bm{e}_1,
\end{equation*}
which gives \(M_{\bm{c}}(x)=\gamma x\) and recovers the CIR
diffusion~\eqref{eq:cir_sde_results}.

The same recovery holds for any linearly independent, centered basis of
polynomial growth whose span contains \(x-\mu\), provided its conjugate
observables are square integrable. The fitted coefficients then represent the
correction \(\gamma(x-\mu)\), which, together with the fixed mean
\(\Phi=\gamma\mu\), yields the exact mobility
(Appendix~\ref{app:cir_benchmark_details}).
\end{revbtext}
\end{revtext}

\subsection{A two-dimensional nonreversible diffusion with affine multiplicative noise}
\label{subsec:affine_multiplicative_2d_results}

\begin{revtext}
\begin{revbtext}
We next consider the two-dimensional diffusion
\begin{equation}
\begin{aligned}
    \d\bm{x}_t
    &=\left[-\nabla U(\bm{x}_t)+\omega\bm{J}\bm{x}_t\right]\d t
      +\bm{\Lambda}(\bm{x}_t)\,\d\bm{W}_t,\\
    U(x,y)&=\tfrac14(x^2+y^2)^2+\tfrac12(x^2+y^2),
\end{aligned}
\label{eq:affine_2d_reference_sde_main}
\end{equation}
where \(\omega=1.1\),
\(\bm J=\bigl(\begin{smallmatrix}0&-1\\1&0\end{smallmatrix}\bigr)\), and
\(\bm\Lambda(x,y)=\bm\Lambda_0+x\bm\Lambda_1+y\bm\Lambda_2\).
The affine noise amplitude gives a quadratic diffusion tensor
\(\bm D_{\mathrm{ref}}=\frac12\bm\Lambda\bm\Lambda^T\), while the rotational
drift breaks detailed balance. This example therefore requires both a
state-dependent diffusion and a circulation.

Our goal is to construct an effective Langevin model from the observed
trajectories, using the stationary score and the derivatives of eighteen target
correlations \(C_{\psi_m,x_n}(t)\), with \(x_1=x\), \(x_2=y\), and
\begin{equation}
    \{\psi_m\}_{m=1}^{9}
    =\{x,y,x^2,xy,y^2,x^3,x^2y,xy^2,y^3\}.
\label{eq:affine_2d_target_observables}
\end{equation}
Unlike in the CIR example, these quantities are estimated from data. We apply
the two calibration procedures of Sec.~\ref{sec:method} to the same trajectory
records. Both calibrations use the same stationary score, estimated by
\revc{implicit score matching with a local energy network (Appendix~\ref{app:stationary_score_network})}.
The system parameters, data, and numerical settings are given in
Appendix~\ref{app:affine_multiplicative_2d_details}.
\end{revbtext}
\end{revtext}

\begin{figure*}[t]
    \centering
    \includegraphics[width=\textwidth]{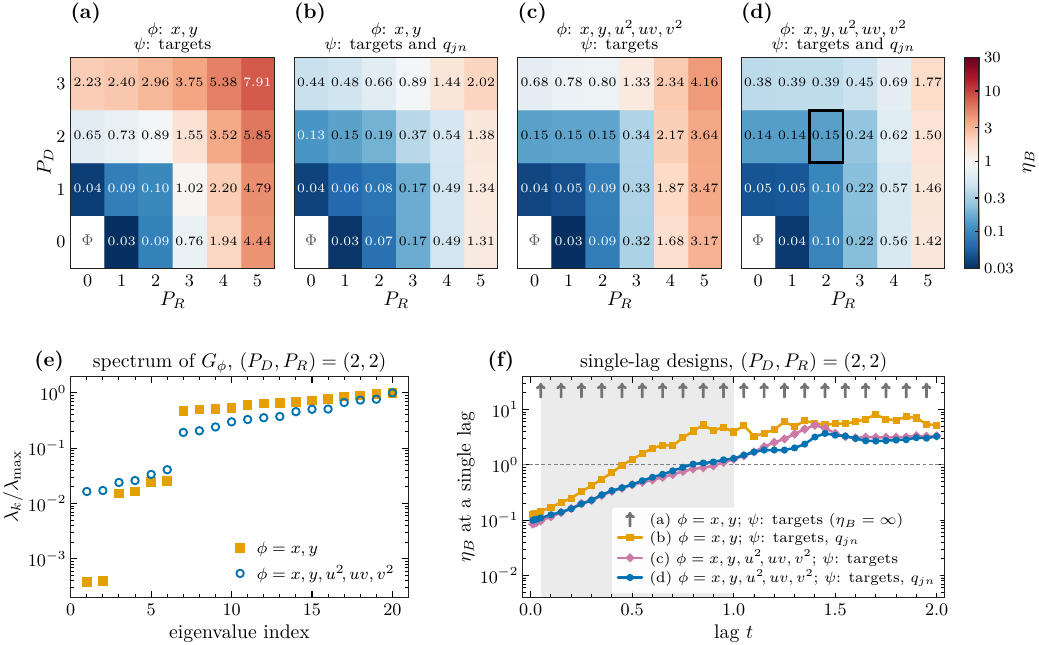}
    \caption{\revb{Choice of observables for the fixed-basis calibration.
    (a--d)~Diagnostic \(\eta_B\), Eq.~\eqref{eq:app_jackknife_eta}, using all
    twenty fitted lags, for polynomial degrees \((P_D,P_R)\) and four choices
    of libraries. ``Targets'' denotes the nine monomials in
    Eq.~\eqref{eq:affine_2d_target_observables}; \(q_{jn}\) are the conjugate
    observables of the largest candidate basis. Blue cells satisfy
    \(\eta_B<1\). The box and panels~(e,f) refer to the mobility basis
    identified by the selection in Fig.~\ref{fig:affine_selection}(a).
    (e)~Eigenvalues of the normalized Gram matrix
    \(\bm G_\phi\), Eq.~\eqref{eq:phi_gram}, for that basis.
    (f)~The same diagnostic using only the equations at each single lag.
    With coordinate initial observables and target future observables, a single
    lag has fewer equations than coefficients, giving
    \(\eta_B=\infty\) (arrows). The gray band marks the fitted lag window.}}
    \label{fig:affine_calibration}
\end{figure*}

\begin{revtext}
\begin{revbtext}
We first perform the fixed-basis calibration. We estimate the mean mobility \(\widehat{\bm\Phi}\) from the initial slope of the
coordinate correlations and expand \(\bm M-\widehat{\bm\Phi}\) on centered
polynomials of the standardized coordinates
\(u=(x-\langle x\rangle)/\sigma_x\) and
\(v=(y-\langle y\rangle)/\sigma_y\). We test candidate mobility bases
that contain these polynomials up to degree \(P_D\) in the symmetric part and
up to degree \(P_R\) in the circulation, for \(0\leq P_D\leq3\) and
\(0\leq P_R\leq5\), and we require the diffusion tensor of each calibrated
model to have no negative eigenvalues on the sampled region. The calibration
equations are evaluated
at twenty lags, \(t=0.05,0.10,\ldots,1.00\), initially using the coordinates
\(\phi_1=x\), \(\phi_2=y\) and the nine target monomials as future observables.

We first choose the observable libraries by comparing the calibration
diagnostics in Fig.~\ref{fig:affine_calibration}(a--d) across the candidate
mobility bases. Coordinates constrain the drift, but some combinations
of diffusion and circulation fields have nearly cancelling contributions to it.
Equation~\eqref{eq:basis_quadratic_response} suggests adding quadratic
observables: their second derivatives constrain the symmetric part directly,
helping distinguish fields with similar adjoint drifts. Comparing
Fig.~\ref{fig:affine_calibration}(a) with~(c), and~(b) with~(d), shows that adding
quadratics allows the coefficients of more candidate mobility bases to be
distinguished despite sampling uncertainty. We therefore add \(u^2\), \(uv\),
and \(v^2\) to the initial library. We enrich the future library with the conjugate observables generated by all 47 fields of the mobility basis with symmetric degree three and circulation degree five. This future library is kept fixed when comparing the candidate bases of the mobility, as prescribed
in Appendix~\ref{app:fixed_basis_details}. The comparison uses the uncertainty
diagnostic \(\eta_B\), for which \(\eta_B<1\) indicates that the fitted equations
distinguish the coefficients above sampling noise
(Appendix~\ref{app:fixed_basis_details}). Adding quadratic
functions to the initial library and conjugate observables to the future library
allows the calibration equations to distinguish the coefficients of nearly every
candidate mobility basis despite sampling uncertainty. The only exception is the
47-field basis, with symmetric degree three and circulation degree five, for which
the uncertainty remains too large to determine all coefficients reliably. We therefore retain
the five initial observables \(x,y,u^2,uv,v^2\) and the targets together with the
conjugate observables at the future time.
\end{revbtext}
\end{revtext}

\begin{figure*}[p]
    \centering
    \includegraphics[width=0.89\textwidth]{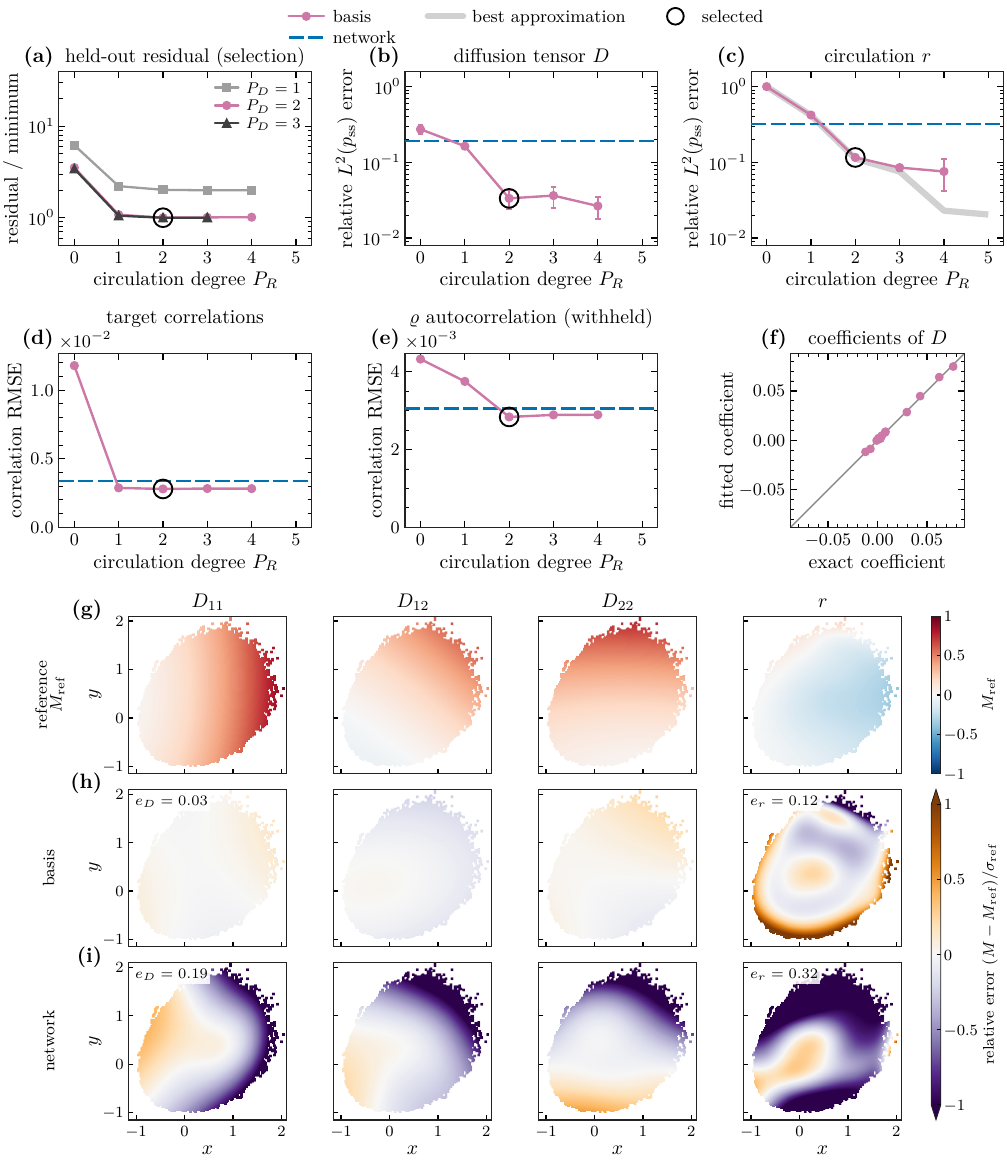}
    \caption{\revb{Field recovery and correlation accuracy for the two
    calibrations. Pink: fixed basis with the stationary score; blue dashed:
    neural mobility with the conditional score. Circles mark the selected basis.
    (a)~Calibration residual on held-out trajectory groups, relative to its
    minimum, for identifiable candidates with \(P_D=1,2,3\).
    (b,c)~Relative field errors \(e_D,e_r\) at \(P_D=2\), with jackknife
    uncertainties; thick gray gives the best polynomial approximation of the
    reference circulation. (d,e)~Simulation errors of the target correlations
    and the squared-amplitude autocorrelation, measured by RMSE using the
    observed variances over the full displayed lag range, \(0\leq t\leq4\).
    (f)~Fitted and exact coefficients of
    the fifteen symmetric basis fields in the selected model.
    (g)~Reference fields on the sampled region.
    (h,i)~Field differences divided by their reference spatial standard
    deviations, for the fixed-basis and neural models.}}
    \label{fig:affine_selection}
\end{figure*}

\begin{revtext}
\begin{revbtext}
With these observable libraries fixed, we next select the mobility basis from the candidates that pass the uncertainty checks. We compare the candidates by fitting their coefficients using one set of trajectories and measuring how well they satisfy the calibration equations on trajectories excluded from that fit.
Figure~\ref{fig:affine_selection}(a) shows a substantial improvement when the
symmetric degree increases from one to two, while further enlargement gives no
statistically resolved gain. We select the smallest basis whose residual
on the excluded trajectories exceeds the minimum over all candidates by less
than one standard error (Appendix~\ref{app:fixed_basis_details}). This basis
has \((P_D,P_R)=(2,2)\) and twenty coefficients, which are then refitted using
all trajectories.

Figure~\ref{fig:affine_calibration}(e,f) provides further diagnostics for this
selected basis. The coordinate Gram matrix is positive definite, yet its smallest
eigenvalues are only \revc{\(4\times10^{-4}\)} of the largest. Adding \(u^2\), \(uv\),
and \(v^2\) raises the smallest relative eigenvalue to \(1.6\times10^{-2}\)
[Fig.~\ref{fig:affine_calibration}(e)]. With coordinates as initial observables, adding the conjugate
observables to the future library lowers \(\eta_B\) from \revc{\(0.89\)} to \revc{\(0.19\)}. Figure~\ref{fig:affine_calibration}(f) shows \(\eta_B\) evaluated with
the equations of a single lag. For the selected libraries, it exceeds one only
near the end of the fitted window, around \(t=1\), where the correlations have
decayed relative to their sampling errors, so that the coefficients are
determined mainly by the equations at shorter lags.

Because the data are generated by a known system, we can also test
whether the basis selected from the data alone in
Fig.~\ref{fig:affine_selection}(a) recovers the mobility of that system. We compare the recovered
diffusion tensor with its exact expression. Since the circulation is not known explicitly, we reconstruct
a reference by fixing the exact diffusion tensor and numerically matching the
model drift to the known drift of the original system. This reconstruction
inherits the estimation error of the stationary score, and the reference fields
are used only for assessment
(Appendix~\ref{app:affine_multiplicative_2d_details}). The relative errors
\(e_D\) and \(e_r\) are root-mean-square field differences normalized by the
stationary spatial variation of each reference. Figure~\ref{fig:affine_selection}(b,c) supports the choice
made in Fig.~\ref{fig:affine_selection}(a): the selected quadratic basis gives \(e_D=\revc{0.033}\) and
\(e_r=\revc{0.116}\). Higher circulation degrees yield no appreciable improvement in the calibrated fields. Figure~\ref{fig:affine_selection}(f) shows that some fitted diffusion coefficients differ from their exact
values by more than the estimated sampling uncertainty. Figure~\ref{fig:affine_selection}(g,h) shows the
reference fields and the corresponding fixed-basis errors, which are largest
near the edge of the sampled region.
More details are provided in
Appendix~\ref{app:affine_multiplicative_2d_details}.
\end{revbtext}
\end{revtext}

\begin{revtext}
\begin{revbtext}
For the second calibration, we use the neural mobility parametrization of
Eq.~\eqref{eq:neural_M_DR_parametrization}, learning both diffusion and
circulation without prescribing a polynomial basis. We fit the eighteen correlations between the nine
observables in Eq.~\eqref{eq:affine_2d_target_observables}, evaluated at the
future time, and the two coordinates, evaluated at the initial time. The conditional score is estimated as described in
Appendix~\ref{app:conditional_score}. Starting from the constant mobility
\(\widehat{\bm\Phi}\) derived before, we train the mobility with the
conditional-score loss~\eqref{eq:deltaM_neural_loss} at thirty lags between
\(0.03\) and \revc{\(1.48\)}.
Figure~\ref{fig:affine_selection}(b,c,i) compares its recovered fields with
those of the fixed-basis model. The neural errors, \(e_D=\revc{0.19}\) and
\(e_r=\revc{0.32}\), are larger, particularly near the edge of the sampled region.
The neural network has many more adjustable parameters than the selected
polynomial model, and different mobility fields may satisfy the fitted
correlation identities with similar accuracy. Errors in the estimated
conditional score also affect the calibration, as quantified in
Appendix~\ref{app:affine_multiplicative_2d_details}.
\end{revbtext}
\end{revtext}

\begin{figure*}[tp]
    \centering
    \includegraphics[width=\textwidth]{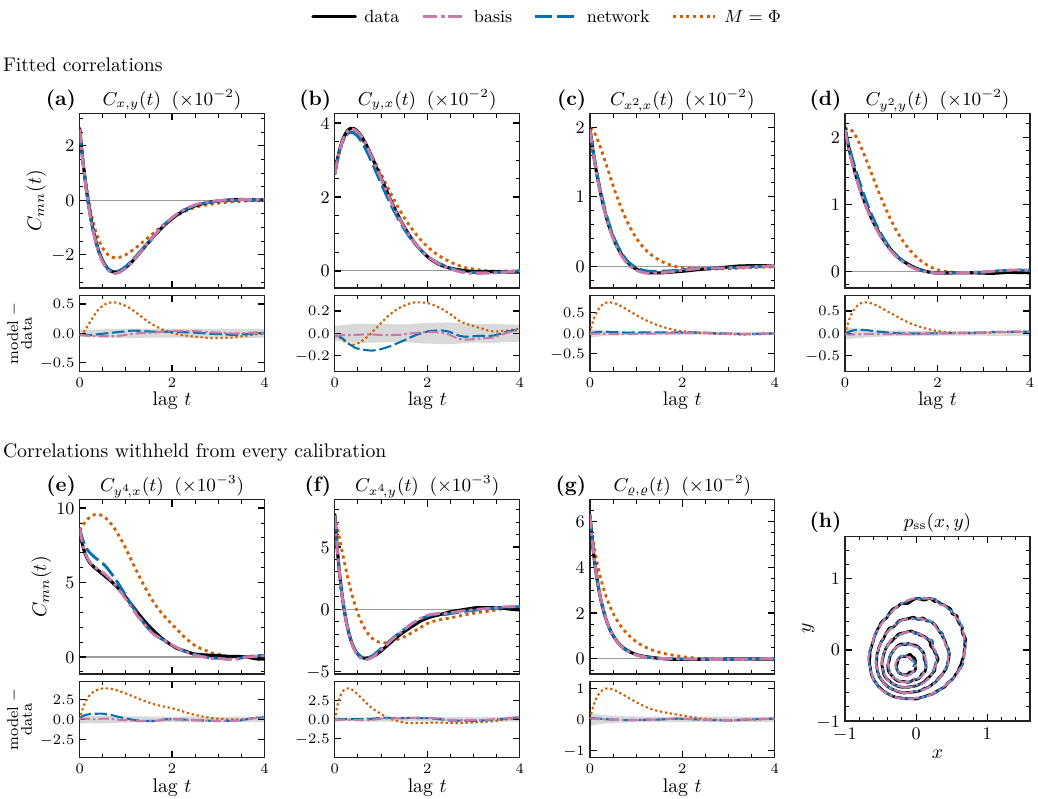}
    \caption{\revb{Correlations and stationary density generated by the
    calibrated two-dimensional models. Each model uses \(1024\) runs with
    \(100\) retained time units, sharing initial states, noise, and the learned
    stationary score. (a--d)~Four of the eighteen target correlations; all are
    shown in Fig.~\ref{fig:affine_forward_all}. (e,f)~Quartic correlations,
    absent from both calibration libraries. (g)~Squared-amplitude
    autocorrelation, an additional diagnostic covered by the span of the
    fixed-basis libraries. Lower strips show model minus data. Gray shading extends to twice the
    combined standard error of the observed and fixed-basis correlation estimates
    on either side of zero. (h)~Stationary-density contours at \(12,30,50,70,90\%\) of
    the maximum observed density.}}
    \label{fig:affine_forward_cphi}
\end{figure*}

\begin{revtext}
\begin{revbtext}
We next integrate both calibrated models and compare the resulting correlations
with the observations. All simulations use the same stationary score, initial
states, and noise. For comparison, we also integrate the Langevin model with the constant
mobility \(\bm M=\widehat{\bm\Phi}\) to test how well the dynamics
can be reproduced using only the estimated mean mobility. Correlations are
computed from fluctuations about the stationary means of the observables.
Their errors are measured by the RMSE between the correlations of each
model and those of the data generated by the true system, over the full displayed lag range,
\(0\leq t\leq4\), after dividing each correlation by the product of
the observed standard deviations of its two observables. The same scale is
used for every model, as detailed in
Appendix~\ref{app:affine_multiplicative_2d_details}.
Besides the eighteen targets, we examine the correlations of the five quartic
monomials with the coordinates and the autocorrelation of the squared amplitude
\(\varrho=x^2+y^2\). The quartic correlations are absent from both calibration
libraries; the squared-amplitude autocorrelation is an additional diagnostic
whose observables lie in the span of the enlarged fixed-basis libraries.

Figure~\ref{fig:affine_forward_cphi} shows that both calibrated models
reproduce the target correlations, the quartic correlations, and the
squared-amplitude autocorrelation much more accurately than the constant
closure. The corresponding errors are given in
Fig.~\ref{fig:affine_selection}(d,e) as functions of the circulation degree of
the fixed basis. They are largest without a circulation and do not decrease
beyond the selected degree, and the neural model gives errors similar to those
of the selected basis. Similar correlation accuracy can therefore coexist with
appreciably different mobility fields. Finally, the nearly identical stationary
densities in Fig.~\ref{fig:affine_forward_cphi}(h) are consistent with the common stationary score; the
improvement over the constant closure is primarily in the dynamics.
\end{revbtext}
\end{revtext}

\begin{figure*}[t]
    \centering
    \includegraphics[width=\textwidth]{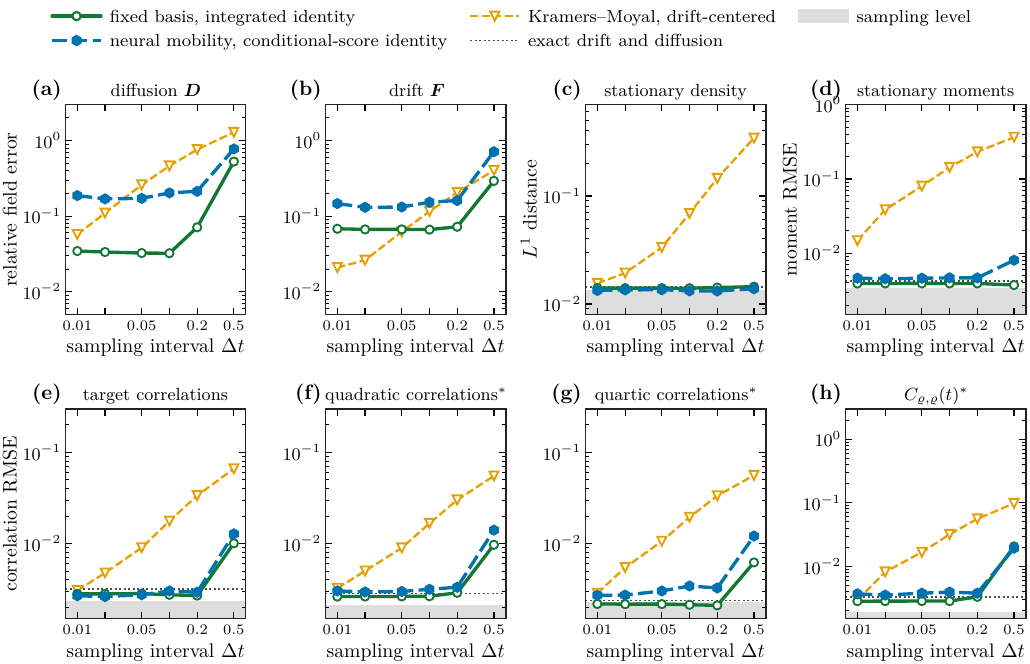}
    \caption{\revc{Calibration of the two-dimensional diffusion from data
    subsampled to the interval \(\Delta t\). Green: fixed basis
    \((P_D,P_R)=(2,2)\) with the integrated identity; blue dashed: neural
    mobility with the integrated conditional-score identity; orange dashed: drift-centered
    Kramers--Moyal estimate with polynomial drift and diffusion libraries of
    \(50\) fields. (a,b)~Relative errors of the diffusion tensor and
    of the drift.
    (c)~\(L^1\) distance of the stationary density. (d)~RMSE of the equal-time
    covariances \(C_{\psi_m,\phi_n}(0)\) of the forward-validation observables,
    scaled as the correlations. (e--h)~RMSE of the target correlations, of the
    correlations of the monomials of degree at most two with the quadratic
    initial observables, of the quartic correlations, and of the
    squared-amplitude autocorrelation. Asterisks mark correlations that are not
    calibration targets; the quadratic and squared-amplitude correlations lie
    in the span of the calibration libraries, whereas the quartic correlations
    lie outside it. Dotted lines: exact system; shading: sampling level.}}
    \label{fig:affine_km}
\end{figure*}

\begin{revctext}
We finally compare the calibrations with Kramers--Moyal (KM) estimates, which
obtain the drift and diffusion from the mean and the second moment of the
increments of the state over one sampling interval, conditioned on the initial
state \cite{Friedrich2011}, as the sampling interval \(\Delta t\) of the
data increases (Fig.~\ref{fig:affine_km}). We use the drift-centered estimator,
which subtracts the fitted drift from the increments before forming the second
moment and thereby removes the squared drift from the diffusion
estimate~\cite{Gottschall2008}. In the KM estimate the drift and the
diffusion tensor are expanded separately in the monomials \(x^ay^b\) with
\(a+b\leq3\), a library of \(50\) fields that contains the exact cubic drift and
quadratic diffusion and is comparable in size to the \(47\) fields of the largest
candidate basis of the fixed-basis calibration. The coefficients of both
expansions follow from the least-squares fits of Appendix~\ref{app:kramers_moyal}.
For each \(\Delta t\), we repeat both calibrations on the subsampled
data with the same stationary and conditional scores, replacing the correlation
derivatives with differences of correlations between consecutive lags through
the integrated identity~\eqref{eq:app_integrated_identity} and its neural
counterpart of Sec.~\ref{subsec:neural_conditional_score}. Only the mean
mobility \(\widehat{\bm\Phi}\) still requires a correlation derivative, the
initial slope of the coordinate correlations. The fixed-basis calibration keeps
the selected \((2,2)\) basis, and the neural calibration adds the quadratic
monomials \(x^2\), \(xy\), and \(y^2\) to its initial observables.

At the finest interval, the fixed basis recovers the diffusion tensor more
accurately than the KM estimate, whereas the KM drift is more accurate, because
the KM library contains the exact drift while the selected basis represents the
circulation only approximately [Fig.~\ref{fig:affine_km}(a,b)]. As \(\Delta t\) grows, the KM
errors increase rapidly, and the simulated KM correlations depart from the
sampling level from \(\Delta t=0.02\). The fixed-basis calibration instead keeps
its field errors nearly unchanged up to \(\Delta t=0.1\), and its simulated
correlations, including those that are not targets, remain at the sampling level
up to \(\Delta t=0.2\), about a fifth of the decorrelation time
[Fig.~\ref{fig:affine_km}(e--h)]. This difference reflects the identities used. KM estimates are exact
only in the limit of a vanishing sampling interval, so their errors grow with
\(\Delta t\). In the integrated identity, the change of each correlation between
two consecutive lags equals an integral over the lag of averages computed from
the data, so no derivative is needed. Its only approximation is the numerical
integration over the lag, which remains accurate as long as the sampling
interval resolves the variation of the correlations. The neural calibration instead has larger field errors but also reproduces the
correlations at the sampling level over the same range.

The stationary statistics separate the two approaches further
[Fig.~\ref{fig:affine_km}(c,d)]. The score-based models share the stationary score, which fixes
the stationary distribution independently of the mobility, so their stationary
densities and equal-time moments remain at the sampling level at every interval.
The KM drift is not tied to the stationary density, and the errors of the fitted
drift and diffusion shift the stationary distribution of the KM model already at
the finest interval, increasingly so as \(\Delta t\) grows.
\end{revctext}

\clearpage

\subsection{Periodic soft-spin stochastic Landau--Lifshitz dynamics}
\label{subsec:periodic_soft_spin_ll_results}

\begin{revtext}
\begin{revbtext}
The third example is a periodic chain of \(N=12\) spins
\(\bm m_i\in\mathbb R^3\) with fluctuating lengths, as in magnetic models with
variable-length moments \cite{Garanin1997,Evans2012,Atxitia2017,MaDudarev2012}.
Writing \(\varrho_i=\|\bm m_i\|^2\), its free energy is
\begin{equation}
\begin{aligned}
 U(\bm m)=\sum_{i=1}^N\Bigl[&
 \tfrac{\lambda}{4}(\varrho_i-m_\ast^2)^2
 +\tfrac J2\|\bm m_{i+1}-\bm m_i\|^2\\
 &-\tfrac{K_z}{2}m_{i,z}^2\Bigr],
\end{aligned}
\label{eq:soft_spin_free_energy_results}
\end{equation}
with periodic indices. The three terms control the spin length, couple
neighboring spins, and favor alignment with the \(z\) axis, respectively
\cite{MaDudarev2012,EllisGalanteSanvito2019}. With effective field
\(\bm h_i=-\nabla_{\bm m_i}U\), the It\^o dynamics are
\begin{equation}
\begin{aligned}
 \d\bm m_i
 &=\left[\bm A_i\bm h_i-\gamma\bm m_i\times\bm h_i\right.\\
 &\quad\left.+\Theta(4\alpha_\parallel-2\alpha_\perp)\bm m_i\right]\d t
 +\sqrt{2\Theta}\,\bm A_i^{1/2}\,\d\bm W_i,\\
 \bm A_i
 &=\varepsilon\bm I+\alpha_\perp
 (\varrho_i\bm I-\bm m_i\bm m_i^T)
 +\alpha_\parallel\bm m_i\bm m_i^T.
\end{aligned}
\label{eq:soft_spin_ll_sde_results}
\end{equation}
The state-dependent damping and noise preserve the Gibbs density
\(p_{\mathrm{ss}}\propto e^{-U/\Theta}\), while precession generates a
probability current along constant-energy surfaces
\cite{Evans2012,MenariniLomakin2020}. In the score-based representation, the
score is \(\bm s_i=\bm h_i/\Theta\) and the mobility is
\begin{equation}
 (\bm M_{\mathrm{true}})_{ij}
 =\delta_{ij}\Theta\bigl(\bm A_i-\gamma[\bm m_i]_\times\bigr),
\label{eq:soft_spin_mobility_results}
\end{equation}
where \([\bm m_i]_\times\bm v=\bm m_i\times\bm v\). Its symmetric part has transverse and longitudinal eigenvalues
\(\lambda_\perp=\Theta(\varepsilon+\alpha_\perp\varrho_i)\) and
\(\lambda_\parallel=\Theta(\varepsilon+\alpha_\parallel\varrho_i)\), which
govern fluctuations in orientation and length. Its antisymmetric part has
precession coefficient \(c_\times=-\Theta\gamma\). The parameters in
Appendix~\ref{app:soft_spin_details} give fast in-plane relaxation and slow
axial relaxation, with decorrelation times of approximately \revc{\(2.9\) and
\(316\)}, respectively.

Our goal is to recover the observed dynamics without assuming the specific functional dependence of the diffusion and precession on the spin configuration.
The targets are the same-site and nearest-neighbor coordinate correlations,
\(N^{-1}\sum_i\langle m_{i+r,d}(t)m_{i,d}(0)\rangle\), with
\(r=0,\pm1\) and \(d=x,y,z\). Rotation about the \(z\) axis and reflection
of the chain reduce them to four distinct curves: in-plane and axial
correlations at the same site and between neighbors. Both calibrations use
the same stationary score, estimated by \revc{implicit score matching with a
local energy network (Appendix~\ref{app:stationary_score_network})}. The data, network settings, and numerical
error definitions are given in Appendix~\ref{app:soft_spin_details}.
\end{revbtext}
\end{revtext}

\begin{figure*}[t]
 \centering
 \includegraphics[width=\textwidth]{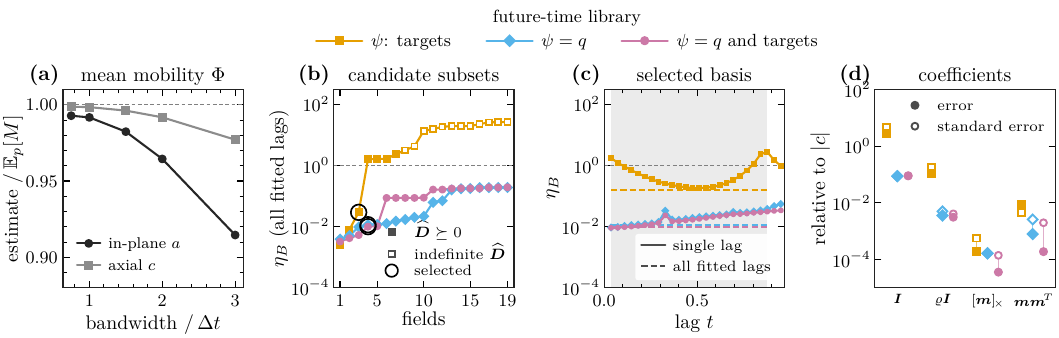}
 \hypersetup{linkcolor=darkred}
 \caption{\revb{Fixed-basis calibration for the soft-spin chain.
 (a)~Mean-mobility components \(a\) and \(c\) in the \(xy\) plane and along
 the \(z\) axis, estimated with local cubic fits and divided by their exact
 values.
 (b)~\(\eta_B\), Eq.~\eqref{eq:app_jackknife_eta}, as fields are progressively
 removed from the mobility library, using all fitted lags. Values below one
 indicate that the equations can distinguish the coefficients despite sampling
 uncertainty. Open symbols mark models whose diffusion tensor has a negative
 eigenvalue within the sampled region; circles mark the subsets selected in
 Fig.~\ref{fig:spin_accuracy}(a).
 (c)~The same quantity for the selected \revc{four-field} basis, computed using
 each lag separately (solid) or all fitted lags together (dashed). Shading
 indicates the lags used for calibration.
 (d)~Errors in the four coefficients corresponding to the true mobility fields
 (filled symbols), compared with their estimated standard errors (open symbols).
 Both are divided by the magnitude of the corresponding exact coefficient.}}
 \label{fig:spin_calibration}
\end{figure*}

\begin{revtext}
\begin{revbtext}
For the fixed-basis calibration, we start from nineteen onsite matrix fields
of polynomial degree at most two, compatible with rotations about the
\(z\) axis. This library contains the true mobility and additional diffusion
and precession fields. We use the expansion of
Eq.~\eqref{eq:basis_ansatz_uncentered}. The coordinate observables
\(\phi_n=m_{i,d}\) suffice: their Gram matrix is positive definite and passes
the uncertainty check, so quadratic initial observables are unnecessary.
For the future observables, we compare the target coordinates, the conjugate
observables \(q_{kn}\) generated by all nineteen mobility fields, and their
union. These observable libraries remain fixed as mobility fields are removed.
The equations use twenty-four lags from \(\Delta t=0.0365\) to \(0.876\).

\revc{The stationary score satisfies the Stein identities to within sampling
uncertainty} (Appendix~\ref{app:soft_spin_details}). The mean mobility is estimated from the initial
slopes of the coordinate correlations, averaged over the twelve sites, and
constrained to have the matrix form required by rotational symmetry about the
\(z\) axis [Eq.~\eqref{eq:soft_spin_axial_class}]. Denoting its in-plane and axial diagonal
entries by \(a\) and \(c\), Fig.~\ref{fig:spin_calibration}(a) shows that both estimates approach a
plateau when the local-cubic bandwidth reaches one sampling interval. The
resulting estimate of \(\bm\Phi\) is used for both the constant closure and
neural calibration, and the same bandwidth is used for the fixed-basis
correlation derivatives. The antisymmetric mean is consistent with zero:
averaging the state-dependent precession removes
it from the constant closure, although it remains important for the dynamics.
\end{revbtext}
\end{revtext}

\begin{figure*}[p]
 \centering
 \includegraphics[width=0.86\textwidth]{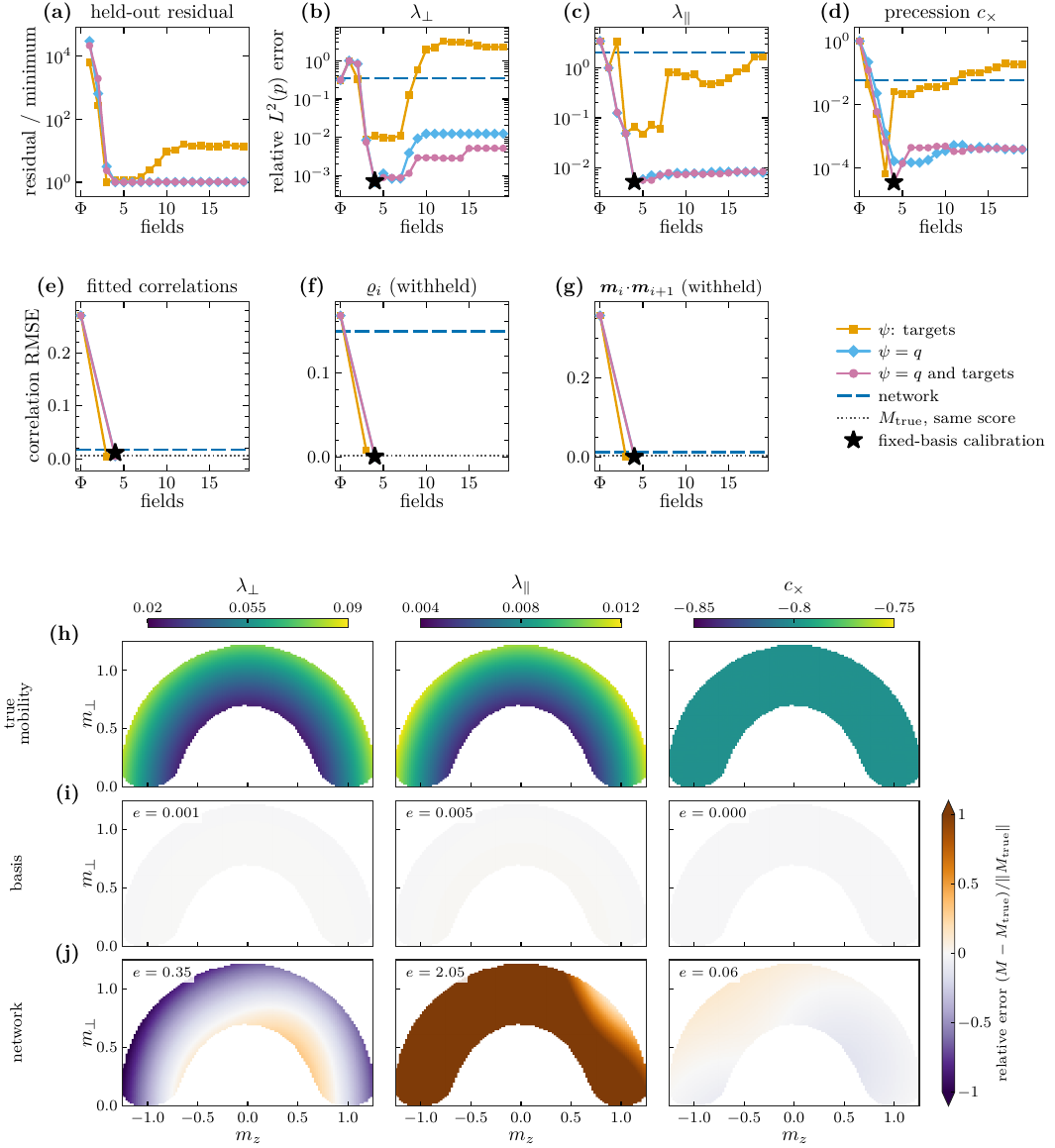}
 \caption{\revb{Mobility selection and validation for the soft-spin chain with
 the network stationary score. Fields are progressively removed from the
 nineteen-field library. Black stars mark the chosen \revc{four-field} mobility used for the reported field maps and forward simulations. (a)~Calibration residual relative to its
 minimum for each future library. (b--d)~Relative local field errors
 \(e_f\), Eq.~\eqref{eq:soft_spin_field_error}. (e--g)~Simulation errors of the four targets
 (averaged), the squared spin length, and the neighboring-spin scalar product,
 measured by RMSE obtained using the observed variances over the full lag ranges in
 Fig.~\ref{fig:spin_forward}. Dotted curves use the true mobility with
 the same stationary score.
 (h)~True local fields as functions of \(m_z\) and
 \(m_\perp=(m_x^2+m_y^2)^{1/2}\). (i,j)~Differences from these fields for the
 fixed-basis and neural models, divided by the respective true-field norms.
 The map construction is specified in Appendix~\ref{app:soft_spin_details}.}}
 \label{fig:spin_accuracy}
\end{figure*}

\begin{figure*}[tp]
 \centering
 \includegraphics[width=\textwidth]{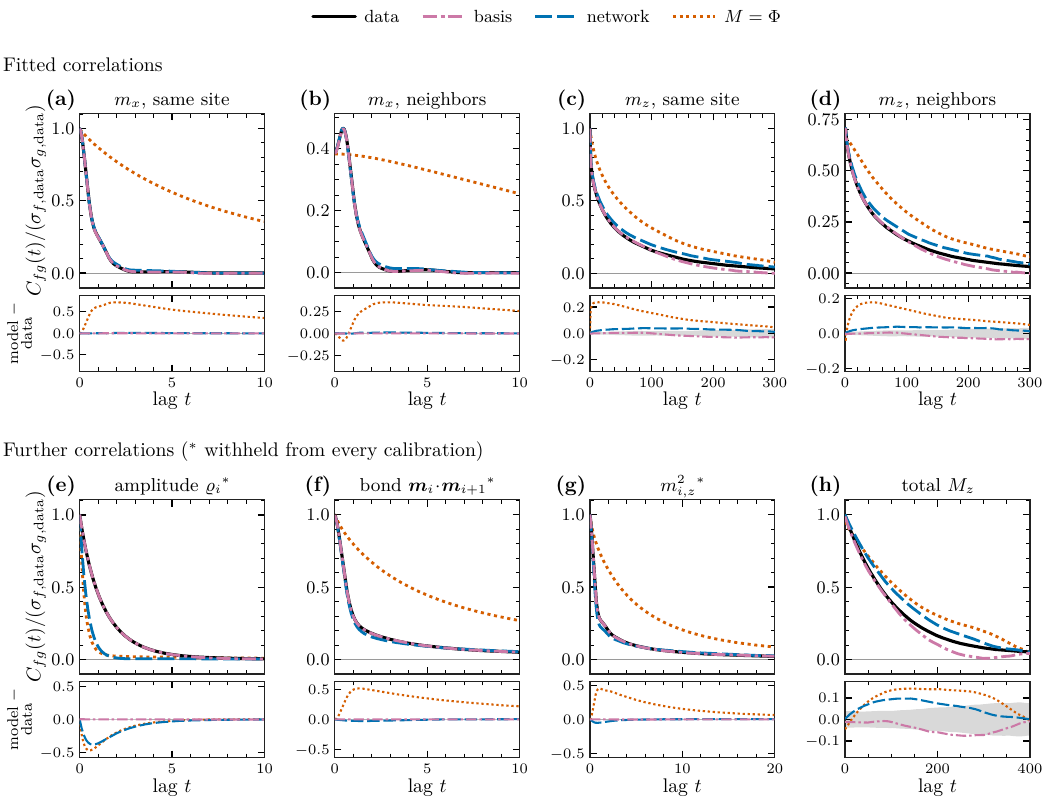}
 \caption{\revb{Correlations generated by the soft-spin models using the same
 network stationary score and \(288\) trajectories of \(500\) retained time
 units. (a--d)~The four target correlations. (e--h)~Additional correlations;
 asterisks mark observables excluded from both calibration libraries.
 Correlations are computed from fluctuations about the stationary means of
 the observables and averaged over equivalent sites. The curves are divided
 by the product of the observed standard deviations of the two observables,
 using the same scale for every model. Lower strips show model minus data.
 Gray shading extends to twice the combined standard error of the data and
 fixed-basis estimates on either side of zero.
 Black: data. Pink dash-dotted: fixed basis. Blue dashed: neural mobility.
 Orange dotted: constant closure \(\bm M=\bm\Phi\).}}
 \label{fig:spin_forward}
\end{figure*}

\begin{revtext}
\begin{revbtext}
{\color{darkred}\hypersetup{linkcolor=darkred}
We progressively reduce the nineteen-field mobility library by removing one
field at a time and comparing the calibration errors of the remaining subsets
on trajectories excluded from the coefficient fit. With the conjugate
observables, every subset satisfies \(\eta_B<1\) and yields a positive
semidefinite diffusion tensor [Fig.~\ref{fig:spin_calibration}(b)]. With the
targets alone, only subsets containing at most \revc{three} fields satisfy this
criterion, and every subset containing \revc{eight} or more fields yields an
indefinite diffusion tensor. Selection using the conjugate observables retains
\revc{four fields, which are exactly the fields of the true mobility}
[Fig.~\ref{fig:spin_accuracy}(a)]. Following the procedure in
Sec.~\ref{subsec:fixed_basis_calibration}, we then refit their coefficients
using both the conjugate observables and the targets.
Figure~\ref{fig:spin_calibration}(c,d) examines this selected basis in more
detail. \revc{With the targets alone, single lags determine its coefficients
within sampling uncertainty only at intermediate lags, whereas the conjugate
observables determine them at every lag [Fig.~\ref{fig:spin_calibration}(c)]. The
fitted coefficients agree with their exact values to within about one standard
error [Fig.~\ref{fig:spin_calibration}(d)].} \par}

For the second calibration, we use a neural mobility with the
positive-definite symmetric part and antisymmetric part described in
Sec.~\ref{subsec:neural_conditional_score}. The same network is applied at
every site, with the local spin and its two neighbors as inputs; its output
specifies the onsite mobility block. The conditional score is \revc{estimated by
implicit score matching with a local energy network}
(Appendix~\ref{app:conditional_score}). Coordinates are used as initial
observables, while the future library contains eleven analytic vector
families, including the target coordinates and nonlinear functions of spin
length, neighboring spins, and axial alignment. They are chosen to have
correlation derivatives that are large relative to sampling uncertainty and
to avoid nearly duplicate curves. We fit the conditional-score constraints over
\(0.256\leq t\leq0.876\), excluding near-zero lags, where sharply
concentrated transition densities make conditional-score estimation difficult.

For forward validation, we integrate each model with the common stationary
score, using \(288\) trajectories of \(500\) retained time units. Besides the
four targets, we examine the autocorrelations of the squared spin length
\(\varrho_i\), the neighboring-spin scalar product
\(\bm m_i\cdot\bm m_{i+1}\), and \(m_{i,z}^2\), all excluded from both
calibrations. The autocorrelation of the total magnetization
\(M_z=N^{-1}\sum_i m_{i,z}\) provides an additional collective diagnostic.
For comparison, we integrate the constant closure and the true mobility with
the same estimated score. Correlation errors are measured by the RMSE of
Eq.~\eqref{eq:app_affine_correlation_rmse}, with all models scaled by the
observed standard deviations, over the full lag range displayed for each
correlation in Fig.~\ref{fig:spin_forward}.

Figure~\ref{fig:spin_accuracy}(b--j) compares the recovered diffusion and
precession fields and the correlations generated by the calibrated models.
We compare diffusion parallel and perpendicular to the spin and the precession
coefficient. Their relative errors are root-mean-square differences from the
true fields, weighted by the observed stationary distribution of spin states
and divided by the corresponding true-field norms
(Appendix~\ref{app:soft_spin_details}). The fixed-basis model gives relative
errors of \revc{\(0.07\%\), \(0.52\%\), and less than \(0.01\%\)} for transverse
diffusion, longitudinal diffusion, and precession [Fig.~\ref{fig:spin_accuracy}(b--d,i)]. The
corresponding neural errors are \revc{\(34\%\), \(204\%\), and \(5.9\%\)}
[Fig.~\ref{fig:spin_accuracy}(b--d,j)]; Fig.~\ref{fig:spin_accuracy}(h) shows the true fields. The smaller precession
errors are consistent with its strong contribution to the fast coordinate
dynamics, whereas the weaker diffusion terms, particularly longitudinal
diffusion, are more sensitive to errors in the estimated \revc{conditional}
score. Despite these differences in the recovered fields, \revc{both models
reproduce the target correlations, with averaged errors of \(0.011\) for the
fixed basis and \(0.018\) for the neural mobility, against \(0.006\) for the true
mobility with the same score} [Fig.~\ref{fig:spin_accuracy}(e)]. Their predictions
differ more clearly for the autocorrelation of the squared spin length,
\(\varrho_i\), which was excluded from calibration and directly probes
fluctuations in spin length. The fixed-basis model reproduces this correlation
\revc{at the level of the true mobility, whereas the neural error is about a
hundred times larger} [Fig.~\ref{fig:spin_accuracy}(f)], consistent with its longitudinal-diffusion
error. \revc{For the autocorrelation of the neighboring-spin scalar product
\(\bm m_i\cdot\bm m_{i+1}\), the fixed-basis error is again at the level of the
true mobility, and the neural error is larger but remains small} [Fig.~\ref{fig:spin_accuracy}(g)].

\end{revbtext}
\end{revtext}

\begin{revtext}
\begin{revbtext}
Figure~\ref{fig:spin_forward} shows that both calibrated models recover the
fast in-plane correlations far more accurately than the constant closure
[Fig.~\ref{fig:spin_forward}(a,b)]. \revc{The slow axial correlations are reproduced by the
fixed-basis model up to a small deficit at lags beyond about \(100\), whereas
the neural model decorrelates too slowly [Fig.~\ref{fig:spin_forward}(c,d)].} The fixed-basis model
captures the squared-length autocorrelation, whereas the neural model makes
it decay much too quickly [Fig.~\ref{fig:spin_forward}(e)], consistent with its error in
longitudinal diffusion. Both models improve the correlations of
\(\bm m_i\cdot\bm m_{i+1}\) and \(m_{i,z}^2\) over the constant closure
[Fig.~\ref{fig:spin_forward}(f,g)], \revc{and both reproduce the slow relaxation of the total
magnetization approximately, the fixed basis decorrelating slightly too fast
and the neural model slightly too slowly} [Fig.~\ref{fig:spin_forward}(h)].
\end{revbtext}
\end{revtext}

\begin{figure*}[t]
 \centering
 \includegraphics[width=\textwidth]{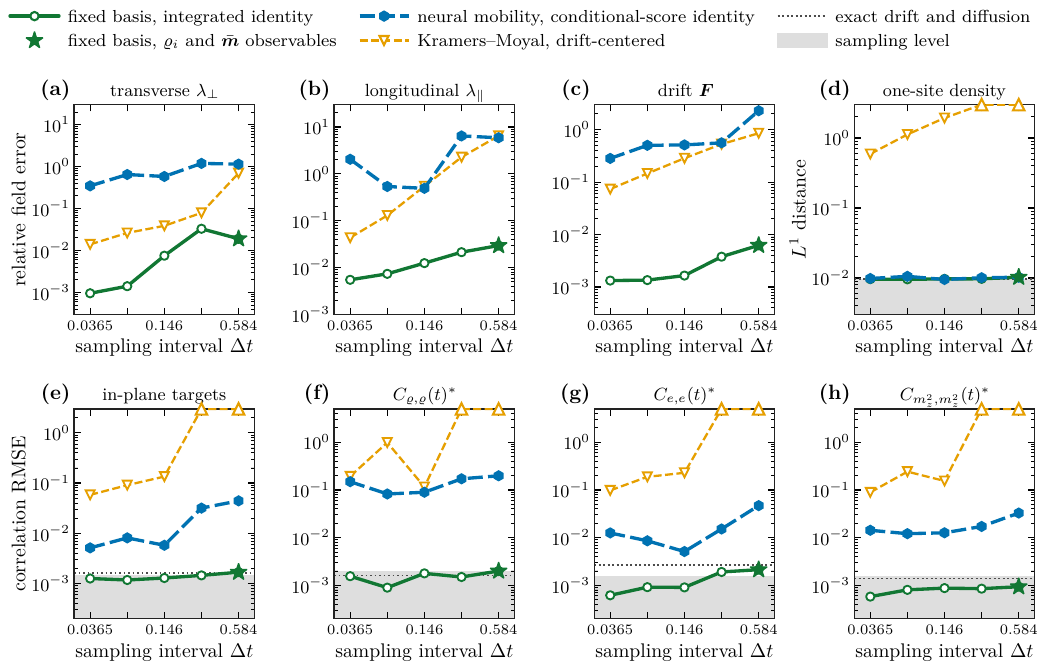}
 \caption{\revc{Calibration of the soft-spin chain from data subsampled to the
 interval \(\Delta t\). Green: four-field fixed basis with the integrated
 identity; stars: the same fields calibrated on the correlations of the spin
 lengths \(\varrho_i\) and of the collective magnetization \(\bar{\bm m}\), which
 remain resolved at \(\Delta t=0.584\); blue dashed: neural mobility with the
 conditional-score identity and local cubic correlation derivatives; orange dashed: drift-centered Kramers--Moyal estimate
 with drift and diffusion libraries of \(34\) fields. (a--c)~Relative
 errors of the
 transverse and
 longitudinal diffusion, Eq.~\eqref{eq:soft_spin_field_error}, and of the drift.
 (d)~\(L^1\) distance of the one-site density of \((|m_z|,m_\perp)\).
 (e--h)~RMSE of the in-plane target correlations, averaged over the same-site
 and neighbor curves, and of the autocorrelations of \(\varrho_i\), of
 \(e_i=\bm m_i\cdot\bm m_{i+1}\), and of \(m_{i,z}^2\), scaled by the observed
 variances; asterisks mark observables excluded from both calibrations. Open
 triangles on the upper edge mark divergent simulations. Dotted lines: true
 mobility with the same score; shading: sampling level.}}
 \label{fig:spin_km}
\end{figure*}

\begin{revctext}
We finally compare the calibrations with drift-centered KM estimates as the
sampling interval increases (Fig.~\ref{fig:spin_km}). In the KM estimate, the drift is expanded in \(23\) vector fields of the spin and of the sum of its
two neighbors, and the diffusion tensor in the \(11\) symmetric onsite fields of
the fixed-basis library, a library of \(34\) fields that contains the exact drift
and diffusion (Appendix~\ref{app:kramers_moyal}). For each \(\Delta t\), we
repeat the score-based calibrations described above on the subsampled data with
the same stationary and conditional scores. The fixed-basis calibration
uses the integrated identity~\eqref{eq:app_integrated_identity}, and since its
expansion does not estimate the mean mobility separately, it requires no
correlation derivative at all. The neural
calibration estimates its correlation derivatives by local cubic fits with
bandwidth \(h=\max(0.11,\Delta t)\).
For the data at their original sampling interval \(\Delta t=0.0365\), the fixed
basis recovers the diffusion about ten times and the drift about fifty times more
accurately than the KM estimate [Fig.~\ref{fig:spin_km}(a--c)]. The correlations generated by the KM model deviate from
the data by more than ten times the sampling uncertainty [Fig.~\ref{fig:spin_km}(e--h)]. These
errors grow with \(\Delta t\), and from \(\Delta t=0.292\) the KM simulations
diverge. The fixed-basis calibration, in contrast, remains accurate up to
\(\Delta t=0.292\), about a tenth of the in-plane decorrelation time, and the
correlations it generates agree with the data within sampling uncertainty over
this range, including those of observables excluded from both calibrations.

At the largest interval, \(\Delta t=0.584\), the data no longer resolve the
fastest spin waves, whose half period is
\(\pi/[\gamma m_\ast(K_z+4J)]\approx0.4\). The coordinate correlations sampled at
this interval therefore misrepresent these oscillations, and the calibration
equations built from them become inconsistent. Two kinds of observables are
largely insensitive to the spin waves: the spin lengths \(\varrho_i\), which precession
leaves unchanged, and the total magnetization
\(\bar{\bm m}=N^{-1}\sum_i\bm m_i\), in which the torques that neighboring spins
exert on each other cancel. At this interval, we therefore calibrate the fixed
basis only on the correlations of \(\varrho_i\) and \(\varrho_{i\pm1}\) with the
initial \(\varrho_i\) and of \(\bar{\bm m}\) with the initial coordinates
\(m_{i,d}\). The resulting model recovers the fields
to within a few percent, keeps the diffusion tensor positive definite, and again
reproduces the correlations within sampling uncertainty [stars]. The neural calibration recovers the precession up to \(\Delta t=0.146\),
but its fields are less accurate than those of KM. Its simulations
nevertheless remain stable and generate correlations closer to the data than
those of KM at every interval. Its fields degrade further from \(\Delta t=0.292\), where the local
cubic fits no longer resolve the fastest spin waves in its correlation
derivatives. Since its accuracy is also limited at every interval by the error
of the estimated conditional score, we apply the change of observables only to
the fixed basis, whose error at \(\Delta t=0.584\) is dominated by aliasing. Finally, as in the two-dimensional example, the
score-based models reproduce the one-site density within sampling uncertainty at
every interval because the stationary score does not depend on the sampling interval, whereas the density of
the KM model differs strongly from the data already at the finest interval
[Fig.~\ref{fig:spin_km}(d)].
\end{revctext}

\clearpage
\makeatletter\global\@colroom\@colht\global\vsize\@colht\makeatother

\section{Conclusions}
\label{sec:conclusions}

\begin{revctext}
In Ref.~\cite{giorgini2026scorelangevin}, we introduced stochastic reduced-order
models whose drift and diffusion are built from the score of the stationary
density and a constant mobility matrix, so that they reproduce by construction
the stationary density and the short-time correlations of the observed
variables. In the present work, we have generalized this framework to
state-dependent mobilities, calibrated through exact identities that relate the
time derivatives of lagged correlation functions to averages over observed pairs
of states and are linear in the mobility. When the mobility is expanded on a
basis of known fields, its coefficients follow from weighted least squares.
Otherwise, the mobility is represented by a neural network and calibrated
through the conditional score of the finite-time transition density. For the
Cox--Ingersoll--Ross process, the fixed-basis calibration recovers the exact
mobility. For the two-dimensional diffusion and the \(36\)-dimensional spin
chain, both calibrations reproduce the finite-lag target correlations much more
accurately than a constant mobility. The fixed basis recovers the diffusion with
errors of \(3\%\) and below \(1\%\), respectively. The neural calibration has
larger field errors, limited by the accuracy of the estimated conditional score
and by the information carried by the selected observables, but it still
reproduces the stationary statistics and the target correlations accurately.

The construction has two practical advantages. First, the stationary density is
fixed by the score independently of the mobility, so every calibrated model
inherits the stationary density of the learned score. Second, the calibration
remains accurate for coarsely sampled data. When the sampling interval resolves
the correlation functions, their time derivatives are estimated directly. When
it does not, the identities are integrated between consecutive lags
(Appendix~\ref{app:fixed_basis_details}), so that no time derivatives of the
correlations are needed. Kramers--Moyal estimates instead require a short
sampling interval. In both examples, they deteriorate rapidly as the interval
grows, whereas the calibration based on the integrated identities remains
accurate (Figs.~\ref{fig:affine_km} and~\ref{fig:spin_km}).

The main limitation of the present study is that the method has been tested
only on fully observed diffusions of the same form as the model, for which a
mobility that reproduces the observed statistics exists. In applications, the
resolved variables are usually projections of a larger system governed by
different dynamics, their evolution is generally non-Markovian, and the
calibrated diffusion acts as an effective closure whose accuracy must be
assessed from the data. This is the setting of the data-driven closures
developed for climate and turbulence models, in which accurate predictions of
the unresolved tendencies do not by themselves guarantee accurate long-time
statistics~\cite{DuraisamyIaccarinoXiao2019,RaspPritchardGentine2018,BoltonZanna2019,ZannaBolton2020,GuillauminZanna2021,Frezat2022}.

Testing the method on such systems is left for future work, but we believe it
has considerable potential in this setting because it builds the stochastic ROM from the
top down. Instead of modeling the unresolved processes and adjusting the model until
its simulations reproduce the data, it starts from the statistical and dynamical
features to be reproduced, constructs the class of models that satisfy them, and
selects a model within this class by imposing physical constraints. The neural calibration shows that
this construction does not require recovering the dynamics that generated the
data. Its mobility differs markedly from the true one, in particular in the
longitudinal diffusion of the spin chain, yet, in these examples, enforcing the weak identities is
sufficient to reproduce the stationary density and the target correlations with
an accuracy comparable to that of the fixed basis. For this reason, we expect
the approach to extend to partially observed systems. With a constant mobility,
this extension has already been carried out for the partially observed Kuramoto--Sivashinsky
equation~\cite{giorgini2026scorelangevin} and for the Planet Simulator climate
model~\cite{giorgini2025cyclostationary}, and preliminary results with a
state-dependent mobility for quasi-geostrophic turbulence are reported in
Ref.~\cite{giorgini2026qg}.
\end{revctext}

\section*{Acknowledgments}

The author thanks Andre N. Souza, Jonah Botvinick-Greenhouse, and Yuanchao Xu for useful discussions.
The author used the AI assistants Claude (Opus 5.5, Anthropic) and Codex
(GPT-6.1, OpenAI) to help revise the text and develop and run the
analysis code. The author directed these
uses, verified all outputs, and takes full responsibility for the content.

\section*{Data Availability}

The Julia \revc{and Python} code and configuration files needed to generate the simulated data,
train the score and mobility models, and reproduce the figures are available in
a public repository~\cite{GiorginiStateDependentMobility2026}.

\clearpage
\onecolumn
\newgeometry{margin=1in}
\appendix
\setcounter{section}{0}
\setcounter{subsection}{0}
\setcounter{equation}{0}
\setcounter{figure}{0}
\setcounter{table}{0}
\renewcommand{\thesection}{S\arabic{section}}
\renewcommand{\thesubsection}{S\arabic{section}.\arabic{subsection}}
\renewcommand{\theequation}{S\arabic{equation}}
\renewcommand{\thefigure}{S\arabic{figure}}
\renewcommand{\thetable}{S\arabic{table}}
\renewcommand{\theHsection}{appendix.\arabic{section}}
\renewcommand{\theHsubsection}{appendix.\arabic{section}.\arabic{subsection}}
\renewcommand{\theHequation}{appendix.\arabic{equation}}
\renewcommand{\theHfigure}{appendix.\arabic{figure}}
\renewcommand{\theHtable}{appendix.\arabic{table}}
\newcommand{\dd}{\,\mathrm{d}}
\newcommand{\R}{\mathbb{R}}
\newcommand{\E}{\mathbb{E}}
\newcommand{\T}{\mathsf{T}}
\newcommand{\Id}{I}
\newcommand{\eps}{\varepsilon}
\newcommand{\meq}{m_{\rm eq}}
\section*{Appendix}
\section{\texorpdfstring{\rev{Exact recovery on a finite observable library}}{Exact recovery on a finite observable library}}
\label{app:finite_library_exact_recovery}

\begin{revtext}
Consider the observable spaces in Eq.~\eqref{eq:observable_family_method}
and a stationary model with the exact observed invariant density \(p\).
Let \(\boldsymbol\psi=(\psi_1,\ldots,\psi_{N_\psi})^T\), and suppose that the observed conditional evolution restricted to \(\mathcal U\) forms a semigroup,
\begin{equation}
K_t^{\mathrm{obs}}\boldsymbol\psi
=
e^{t\bm A}\boldsymbol\psi
\label{eq:neural_observable_semigroup}
\end{equation}
for a fixed matrix \(\bm A\). Define the generator residual
\begin{equation}
\bm r_\theta
=
\mathcal L_{\bm M_\theta}\boldsymbol\psi
-\bm A\boldsymbol\psi.
\label{eq:neural_generator_residual}
\end{equation}
By Eqs.~\eqref{eq:weak_generator_identity} and~\eqref{eq:neural_observable_semigroup}, the weak calibration residual at lag \(t\) satisfies
\begin{equation}
\begin{aligned}
R^\theta_{mn}(t)
&:=\left\langle
\psi_m(\bm x_t)
(\mathcal L_{\bm M_\theta}^{\dagger}\phi_n)(\bm x_0)
\right\rangle_{\mathrm{obs}}
-\dot C_{mn,\mathrm{obs}}(t)
\\
&=\sum_{j=1}^{N_\psi}
(e^{t\bm A})_{mj}\left\langle r_{\theta,j}\phi_n\right\rangle_p.
\end{aligned}
\label{eq:app_neural_residual_transport}
\end{equation}
At the population level, with the exact stationary score, enforcing the weak identities for all library pairs at a common lag sets \(R^\theta_{mn}(t)=0\). Since \(e^{t\bm A}\) is invertible, it follows that each component of \(\bm r_\theta\) is orthogonal to \(\mathcal V\):
\begin{equation}
\left\langle r_{\theta,m}\phi_n\right\rangle_p=0.
\label{eq:neural_residual_orthogonality}
\end{equation}
If, in addition, each component \(r_{\theta,m}\) belongs to \(\mathcal V\), Eq.~\eqref{eq:neural_residual_orthogonality} with \(\phi_n=r_{\theta,m}\) gives \(\langle r_{\theta,m}^2\rangle_p=0\), so that \(\bm r_\theta=0\) and \(\mathcal L_{\bm M_\theta}\boldsymbol\psi=\bm A\boldsymbol\psi\). The conditional means of the model then obey
\begin{equation}
\partial_t K_t^{\bm M_\theta}\boldsymbol\psi
=K_t^{\bm M_\theta}\mathcal L_{\bm M_\theta}\boldsymbol\psi
=\bm A\,K_t^{\bm M_\theta}\boldsymbol\psi,
\qquad
K_t^{\bm M_\theta}\boldsymbol\psi=e^{t\bm A}\boldsymbol\psi=K_t^{\mathrm{obs}}\boldsymbol\psi,
\end{equation}
which coincide with the observed ones at every lag. Since the model also has the observed stationary density \(p\), its correlations
\(C_{\psi_m,g}(t)=\langle(K_t^{\bm M_\theta}\psi_m)\,g\rangle_p\) equal the observed ones for every \(g\), including the targets. A single lag fitted exactly thus determines the correlations of \(\mathcal U\) at all lags.
\end{revtext}

\section{\texorpdfstring{Implementation of the calibration}{Implementation of the calibration}}
\label{app:fixed_basis_details}
\label{app:local_polynomial_cdot_estimation}

\begin{revctext}
Population averages under the observed stationary process are denoted by
\(\langle\cdot\rangle\) and empirical estimates by hats. The rows of
Eq.~\eqref{eq:library_linear_system} are indexed by \(\alpha=(\ell,m,n)\), and the
weighted system is
\begin{equation}
    \bm B_w=\bm W^{1/2}\bm B,
    \qquad
    \bm y_w=\bm W^{1/2}\bm y .
\end{equation}

\paragraph*{Centered expansion.}
In the CIR and two-dimensional examples, the mean
mobility~\eqref{eq:Phi_from_Cdot_zero_method} is held fixed,
\begin{equation}
    \bm M_{\bm c}=\bm\Phi+\sum_{k=1}^{K}c_k\bm H_k,
    \qquad
    \langle\bm H_k\rangle=0,
    \qquad
    y_\alpha=\dot C_{mn,\mathrm{obs}}(t_\ell)
      -\bigl\langle\psi_m(\bm x_{t_\ell})\,
       (\mathcal L^\dagger_{\bm\Phi}\phi_n)(\bm x_0)\bigr\rangle_{\mathrm{obs}} .
    \label{eq:app_centered_expansion}
\end{equation}
For the uncentered expansion~\eqref{eq:basis_ansatz_uncentered}, used for
the soft-spin chain, the fixed term \(\bm\Phi\) and the condition
\(\langle\bm H_k\rangle=0\) are omitted, here and in the formulas below.

\paragraph*{Relative diagnostics.}
The diagnostics test whether \(\bm G=\bm G_\phi\succ0\) and the full column
rank of \(\bm B_w\) persist under sampling errors. With
\(\bm E=\widehat{\bm G}^{-1/2}(\bm G-\widehat{\bm G})\widehat{\bm G}^{-1/2}\),
\(\bm\Delta=\bm B_w-\widehat{\bm B}_w\), and the pseudoinverse
\(\widehat{\bm B}_w^{+}\), for which
\(\widehat{\bm B}_w^{+}\widehat{\bm B}_w=\bm I\),
\begin{equation}
\begin{aligned}
    \bm G&=\widehat{\bm G}^{1/2}(\bm I+\bm E)\,\widehat{\bm G}^{1/2},
    &\qquad
    \|\bm E\|_2<1\ &\Longrightarrow\ \bm G\succ0,\\
    \bm B_w\bm v&=\bigl(\bm I+\bm\Delta\widehat{\bm B}_w^{+}\bigr)\widehat{\bm B}_w\bm v,
    &\qquad
    \|\bm\Delta\widehat{\bm B}_w^{+}\|_2<1\ &\Longrightarrow\ \operatorname{rank}\bm B_w=K .
\end{aligned}
\end{equation}
Both relative perturbations are invariant under
\(\bm G\mapsto\bm T^T\bm G\bm T\) and \(\bm B\mapsto\bm B\bm T\) for any
invertible \(\bm T\), and hence under any rescaling or linear recombination of
the coefficient directions. They are estimated from the replicates
\(\widehat{\bm G}_{(j)}\) and \(\widehat{\bm B}_{w,(j)}\), computed with trajectory
\(j\) of the \(N_{\mathrm{tr}}\) trajectories omitted and with \(\bm W\) fixed,
whose means are \(\overline{\bm G}\) and \(\overline{\bm B}_w\),
\begin{equation}
\begin{aligned}
    \widehat{\eta}_G
    &=
    \Bigl[\frac{N_{\mathrm{tr}}-1}{N_{\mathrm{tr}}}\sum_{j=1}^{N_{\mathrm{tr}}}
    \bigl\|\widehat{\bm{G}}^{-1/2}(\widehat{\bm{G}}_{(j)}-\overline{\bm{G}})
    \widehat{\bm{G}}^{-1/2}\bigr\|_2^2\Bigr]^{1/2},
    \\
    \widehat{\eta}_B
    &=
    \Bigl[\frac{N_{\mathrm{tr}}-1}{N_{\mathrm{tr}}}\sum_{j=1}^{N_{\mathrm{tr}}}
    \bigl\|(\widehat{\bm{B}}_{w,(j)}-\overline{\bm{B}}_w)
    \widehat{\bm{B}}_w^{+}\bigr\|_2^2\Bigr]^{1/2}.
\end{aligned}
    \label{eq:app_jackknife_eta}
\end{equation}
Values below one indicate that positive definiteness and full column rank are
supported by the data.

\paragraph*{Positivity constraint and coefficient errors.}
The calibrated model is a diffusion only if its diffusion tensor
\(\bm D_{\bm c}=\operatorname{sym}\bm M_{\bm c}\) is positive semidefinite, which
the least-squares fit~\eqref{eq:basis_wls} does not guarantee. We therefore
require
\begin{equation}
    \bm D_{\bm c}(\bm x_g)
    =\operatorname{sym}\bm\Phi+\sum_{k=1}^{K}c_k\operatorname{sym}\bm H_k(\bm x_g)
    \succeq0
    \label{eq:app_positivity_grid}
\end{equation}
at the points \(\bm x_g\) of a grid covering the region visited by the data.
Each constraint states that a matrix depending linearly on \(\bm c\) is positive
semidefinite. For a \(2\times2\) matrix with entries \(a\), \(b\), and \(d\), this
is equivalent to the second-order-cone condition
\begin{equation}
    a+d\geq\bigl\|(a-d,\,2b)\bigr\|_2 ,
\end{equation}
and for larger matrices it is a semidefinite constraint. Since the weighted
least-squares objective is a convex quadratic function of \(\bm c\) and every
constraint defines a convex set of coefficients, the constrained problem is
convex, and its minimum is computed with an interior-point conic solver. The
standard errors of the coefficients are obtained by repeating the whole
calibration with each trajectory omitted in turn, including the constrained fit
and, for the centered expansion, the estimation of \(\bm\Phi\),
\begin{equation}
    \sigma_{c_k}^2
    =\frac{N_{\mathrm{tr}}-1}{N_{\mathrm{tr}}}\sum_{j=1}^{N_{\mathrm{tr}}}
     \bigl(\widehat c_{k,(j)}-\overline c_k\bigr)^2 .
\end{equation}

\paragraph*{Correlation derivatives.}
The calibration equations involve the derivatives \(\dot{\bm C}_{m,n}(\tau)\)
of empirical lagged correlations. Because finite differences would amplify the
sampling fluctuations of these correlations by the reciprocal of the spacing between
lags, the derivatives are obtained from local polynomial fits in the lag to
the empirical correlations
\begin{equation}
    \widehat{\bm C}_{m,n}(\tau_\ell)
    =\frac1{N_\ell}\sum_{(r,s)\in\mathcal P_\ell}
    \psi_m(\bm x(t_s))\,\phi_n(\bm x(t_r))^T,
\end{equation}
where \(\mathcal P_\ell\) contains the \(N_\ell\) index pairs with
\(t_s-t_r=\tau_\ell\). For a target lag \(\tau_\ast\) and a
bandwidth \(h\), each scalar entry \(\widehat C_\ell\) of
\(\widehat{\bm C}_{m,n}(\tau_\ell)\) is fitted by a cubic polynomial whose
coefficients minimize
\begin{equation}
    \sum_\ell K\!\left(\frac{\tau_\ell-\tau_\ast}{h}\right)
    \left[\widehat C_\ell-\sum_{q=0}^{3}\frac{\alpha_q}{q!}
    \left(\frac{\tau_\ell-\tau_\ast}{h}\right)^q\right]^2 ,
\end{equation}
and the derivative estimate at \(\tau_\ast\) is \(\widehat\alpha_1/h\). The kernel is
the truncated Gaussian \(K(v)=e^{-v^2/2}\) for \(|v|\leq4\) and \(K(v)=0\)
otherwise. To avoid an endpoint bias, the support lags extend beyond the largest
fitted lag, and at zero lag the fitting window is one-sided. The bandwidth must
resolve the fastest time scale of the correlation curves. It is selected from the
data as the largest value below which the derivative estimate no longer changes,
and the values used in the examples are given in Appendix~\ref{app:examples_details}.
When the sampling interval is too coarse for such a bandwidth, the
integrated form of the next paragraph is preferable.
Since the derivative estimate is a weighted sum of the empirical correlations,
with weights fixed by the lags, the kernel, and the bandwidth, the jackknife
replicates defined above are obtained by applying
the same weights to the correlations computed with one trajectory omitted.

\paragraph*{Integrated form of the identity.}
Integrating Eq.~\eqref{eq:library_linear_system}, including the contribution of
the mean, over \([t_{\ell-1},t_\ell]\) avoids differentiating the correlations,
\begin{equation}
    C_{mn,\mathrm{obs}}(t_\ell)-C_{mn,\mathrm{obs}}(t_{\ell-1})
    \approx
    \int_{t_{\ell-1}}^{t_\ell}
    \Bigl[\sum_kB_{mn,k}(t)\,c_k
    +\bigl\langle\psi_m(\bm{x}_t)\,
    (\mathcal{L}^\dagger_{\bm{\Phi}}\phi_n)(\bm{x}_0)\bigr\rangle\Bigr]\d t,
    \label{eq:app_integrated_identity}
\end{equation}
where \(B_{mn,k}(t)=\langle\psi_m(\bm x_t)\,q_{kn}(\bm x_0)\rangle\). Its
left-hand side is a difference of two empirical correlations, so, unlike the
derivative form, it requires neither a derivative estimate nor the choice of a
bandwidth. The integrals are evaluated on the
saved lags by four-point Lagrange quadrature,
\begin{equation}
    \int_{t_{\ell-1}}^{t_\ell}f\,\d t
    \simeq
    \begin{cases}
      \dfrac{\Delta t}{24}\bigl(-f_{\ell-2}+13f_{\ell-1}+13f_\ell-f_{\ell+1}\bigr),
        & \ell\geq2,\\[1.5ex]
      \dfrac{\Delta t}{24}\bigl(9f_0+19f_1-5f_2+f_3\bigr),
        & \ell=1,
    \end{cases}
    \qquad f_j=f(t_j),
\end{equation}
with a local error of order \(\Delta t^5\). The quadrature involves only
the four lags around each interval, whereas each local-polynomial derivative
estimate extends over several bandwidths, and the bandwidth cannot be smaller
than the sampling interval. The integrated form therefore tolerates coarser
sampling.

\paragraph*{Basis selection.}
A candidate basis is a subset \(\mathcal S\subseteq\{1,\dots,K\}\) of the
fields of a common family, with mobility
\(\bm M_{\bm c}=\bm\Phi+\sum_{k\in\mathcal S}c_k\bm H_k\). Its weighted
calibration system is \(\bm B_{w,\mathcal S}\bm c\approx\bm y_w\), where
\(\bm B_{w,\mathcal S}\) contains the columns \(k\in\mathcal S\) of \(\bm B_w\).
The admissible candidates are
\begin{equation}
    \mathcal C=\bigl\{\mathcal S:\ \widehat\eta_G(\mathcal S)<1,\
      \widehat\eta_B(\mathcal S)<1\bigr\},
\end{equation}
with the diagnostics~\eqref{eq:app_jackknife_eta}. The trajectories are
partitioned into disjoint groups \(\mathcal T_1,\dots,\mathcal T_{N_f}\). Let
\(\widehat{\bm B}^{(f)}_{w,\mathcal S}\) and \(\widehat{\bm y}^{(f)}_w\) be the
weighted conjugate-observable equations estimated from the trajectories in
\(\mathcal T_f\), and \(\widehat{\bm B}^{(-f)}_{w,\mathcal S}\) and
\(\widehat{\bm y}^{(-f)}_w\) those estimated from all other trajectories, with
the same weights \(\bm W\). For each candidate and group,
\begin{equation}
\begin{aligned}
    \widehat{\bm c}^{(-f)}_{\mathcal S}
    &=\operatorname*{arg\,min}_{\bm c}
      \bigl\|\widehat{\bm B}^{(-f)}_{w,\mathcal S}\bm c-\widehat{\bm y}^{(-f)}_w\bigr\|_2^2,
    \\
    \rho_f(\mathcal S)
    &=\bigl\|\widehat{\bm B}^{(f)}_{w,\mathcal S}\,\widehat{\bm c}^{(-f)}_{\mathcal S}
      -\widehat{\bm y}^{(f)}_w\bigr\|_2^2,
    \qquad
    \bar\rho(\mathcal S)=\frac1{N_f}\sum_{f=1}^{N_f}\rho_f(\mathcal S),
\end{aligned}
\end{equation}
so that \(\bar\rho(\mathcal S)\) is the residual on trajectories not used in the
fit. The fit of each group is subject to the positivity
constraint~\eqref{eq:app_positivity_grid} whenever the fit to all trajectories
requires it. With \(\mathcal S^\ast=\operatorname*{arg\,min}_{\mathcal S\in\mathcal C}\bar\rho(\mathcal S)\)
and \(\Delta_f(\mathcal S)=\rho_f(\mathcal S)-\rho_f(\mathcal S^\ast)\), the
selected basis is the smallest candidate within one standard error of
\(\mathcal S^\ast\),
\begin{equation}
\begin{aligned}
    \widehat{\mathcal S}
    &=\operatorname*{arg\,min}_{\mathcal S\in\mathcal C}\bigl\{|\mathcal S|:\
      \bar\Delta(\mathcal S)\leq\sigma_\Delta(\mathcal S)\bigr\},
    \\
    \bar\Delta(\mathcal S)&=\frac1{N_f}\sum_{f=1}^{N_f}\Delta_f(\mathcal S),
    \qquad
    \sigma_\Delta^2(\mathcal S)=\frac{1}{N_f(N_f-1)}
      \sum_{f=1}^{N_f}\bigl[\Delta_f(\mathcal S)-\bar\Delta(\mathcal S)\bigr]^2 .
\end{aligned}
\end{equation}
The coefficients of \(\widehat{\mathcal S}\) are then refitted to the equations of
all libraries using all trajectories.
\end{revctext}

\section{\texorpdfstring{\revc{Score estimation by implicit score matching}}{Score estimation by implicit score matching}}
\label{app:score_estimation}

\begin{revctext}
The stationary score \(\bm s(\bm x)=\nabla_{\bm x}\log p_{\mathrm{ss}}(\bm x)\) and
the conditional score
\(\bm s_{t|0}(\bm x_t\mid\bm x_0)=\nabla_{\bm x_0}\log p_t(\bm x_t\mid\bm x_0)\)
are estimated from the trajectory data by implicit score
matching~\cite{Hyvarinen2005}. We approximate \(\bm s\) by a network
\(\bm s_\theta\) with parameters \(\theta\)
(Appendix~\ref{app:stationary_score_network}) and minimize
\begin{equation}
    \mathcal{J}_{\mathrm{ISM}}(\theta)
    =
    \Bigl\langle
        \tfrac12\|\bm{s}_{\theta}\|^2
        +
        \nabla\cdot\bm{s}_{\theta}
    \Bigr\rangle_{p_{\mathrm{ss}}}.
    \label{eq:app_ism_objective}
\end{equation}
Because \(p_{\mathrm{ss}}\bm s=\nabla p_{\mathrm{ss}}\), integration by parts gives
\begin{equation}
    \langle\bm s_\theta\cdot\bm s\rangle_{p_{\mathrm{ss}}}
    =\int\bm s_\theta\cdot\nabla p_{\mathrm{ss}}\,\d\bm x
    =-\langle\nabla\cdot\bm s_\theta\rangle_{p_{\mathrm{ss}}},
\end{equation}
provided that \(p_{\mathrm{ss}}\bm s_\theta\) vanishes on the boundary of the
state space. Expanding the square then yields
\begin{equation}
\begin{aligned}
    \tfrac12\bigl\langle\|\bm s_\theta-\bm s\|^2\bigr\rangle_{p_{\mathrm{ss}}}
    &=\tfrac12\langle\|\bm s_\theta\|^2\rangle_{p_{\mathrm{ss}}}
     -\langle\bm s_\theta\cdot\bm s\rangle_{p_{\mathrm{ss}}}
     +\tfrac12\langle\|\bm s\|^2\rangle_{p_{\mathrm{ss}}}\\
    &=\mathcal J_{\mathrm{ISM}}(\theta)
     +\tfrac12\langle\|\bm s\|^2\rangle_{p_{\mathrm{ss}}}.
\end{aligned}
\end{equation}
The last term does not depend on \(\theta\), so minimizing
\(\mathcal J_{\mathrm{ISM}}\), which requires only samples of \(p_{\mathrm{ss}}\),
minimizes the mean squared error of \(\bm s_\theta\).
Unlike denoising score matching~\cite{Vincent2011}, this objective
involves no noise amplitude, so the estimated score is not smoothed. It is
well suited to stochastic systems, whose stationary density is smooth, such as
the examples considered here. For deterministic chaotic systems, the invariant
measure can be concentrated on a fractal attractor, and the score must then be
regularized by adding noise, as in denoising score matching.
\end{revctext}

\subsection{\texorpdfstring{\revc{Stationary score with local energy networks}}{Stationary score with local energy networks}}
\label{app:stationary_score_network}

\begin{revctext}
We use the parametric log-density
\begin{equation}
    \log p_\theta(\bm x)
    =\sum_{i=1}^{N}u_\theta(\bm\xi_i)-c_\theta(\bm x)+\mathrm{const},
    \qquad
    \bm\xi_i=\bm\xi(\bm x_{\mathcal N_i}),
    \qquad
    \bm s_\theta=\nabla_{\bm x}\log p_\theta,
    \label{eq:app_local_energy}
\end{equation}
where \(\bm x_{\mathcal N_i}\) collects the coordinates of the neighborhood
\(\mathcal N_i\) of site \(i\), formed by the \(P\) sites within a distance \(r\)
of it, \(\bm\xi\) is a fixed feature map with components standardized over the
data, and \(c_\theta\) is a confining term. The local
energy \(u_\theta\) is a residual network with \(L\) blocks, SiLU activations
\(\sigma(z)=z/(1+e^{-z})\), and a linear readout of the features,
\begin{equation}
\begin{aligned}
    \bm h^{(0)}&=\sigma\bigl(\bm W^{(0)}\bm\xi+\bm b^{(0)}\bigr),\\
    \bm h^{(\ell)}&=\bm h^{(\ell-1)}
      +L^{-1/2}\,\sigma\bigl(\bm W^{(\ell)}\bm h^{(\ell-1)}+\bm b^{(\ell)}\bigr),
      \qquad\ell=1,\dots,L,\\
    u_\theta(\bm\xi)&=\bm w_{\mathrm{out}}^T\bm h^{(L)}+\bm w^T\bm\xi .
\end{aligned}
\label{eq:app_residual_network}
\end{equation}
Since \(u_\theta(\bm\xi_i)\) depends only on \(\bm x_{\mathcal N_i}\), the divergence
in Eq.~\eqref{eq:app_ism_objective} is a sum of local Laplacians,
\begin{equation}
    \nabla\cdot\bm s_\theta
    =\sum_{i=1}^{N}\Delta_{\mathcal N_i}u_\theta(\bm\xi_i)-\Delta c_\theta,
\end{equation}
where \(\Delta_{\mathcal N_i}\) acts on \(\bm x_{\mathcal N_i}\). For each
component \(z_k\) of a pre-activation in Eq.~\eqref{eq:app_residual_network},
\begin{equation}
    \nabla\sigma(z_k)=\sigma'(z_k)\,\nabla z_k,
    \qquad
    \Delta\sigma(z_k)=\sigma'(z_k)\,\Delta z_k+\sigma''(z_k)\,|\nabla z_k|^2 ,
    \label{eq:app_forward_laplacian}
\end{equation}
and these rules are applied layer by layer, starting from \(\nabla\bm\xi\) and
\(\Delta\bm\xi\). The Laplacian is therefore computed exactly.
The network is trained with Adam and an exponential moving average of the
weights.

For the two-dimensional diffusion,
\begin{equation}
    N=1,\qquad
    \bm\xi=\tilde{\bm x},\qquad
    c_\theta=\tfrac12\,\tilde{\bm x}^T\bm K_\theta\tilde{\bm x},\qquad
    \bm K_\theta\succ0,\qquad
    L=3,\qquad
    \dim\bm h^{(\ell)}=64,
\end{equation}
where \(\tilde{\bm x}\) denotes the standardized coordinates, and all saved
states after burn-in are used.

For the soft-spin chain, \(r=1\), so that \(P=3\), \(L=3\), and
\(\dim\bm h^{(\ell)}=128\). The features of a neighborhood with spins
\(\bm m_1,\dots,\bm m_P\) are their quadratic invariants under rotations about
the \(z\) axis and under \(\bm m\to-\bm m\),
\begin{equation}
    \bm\xi=\Bigl(\{m_{p,z}m_{q,z}\}_{p\leq q},\;
      \{\bm m_p^\perp\cdot\bm m_q^\perp\}_{p\leq q},\;
      \{(\bm m_p\times\bm m_q)_z\}_{p<q}\Bigr),
    \qquad
    \bm m^\perp=(m_x,m_y),
    \label{eq:app_spin_invariants}
\end{equation}
The reflection symmetry of the chain and the confinement are imposed by
\begin{equation}
    u_\theta(\bm\xi_i)\;\to\;\tfrac12\bigl[u_\theta(\bm\xi_i)+u_\theta(\bm\xi_i^{R})\bigr],
    \qquad
    c_\theta=\tfrac14\kappa\sum_{i=1}^{N}\varrho_i^2,
    \qquad
    \kappa>0,
\end{equation}
where \(\bm\xi_i^{R}\) is evaluated with the sites of \(\mathcal N_i\) in reversed
order. For
\(\bm w_{\mathrm{out}}=0\), the score is linear in \(\bm\vartheta=(\bm w,\kappa)\),
\begin{equation}
    \bm s_\theta=\bm\Xi\,\bm\vartheta,
    \qquad
    \bm\Xi=\Bigl[\,\tfrac12\nabla_{\bm x}\sum_{i=1}^{N}\bigl(\bm\xi_i+\bm\xi_i^{R}\bigr)^T,\;
      -\tfrac14\nabla_{\bm x}\sum_{i=1}^{N}\varrho_i^2\,\Bigr],
\end{equation}
so that Eq.~\eqref{eq:app_ism_objective} is quadratic in \(\bm\vartheta\),
\begin{equation}
    \mathcal J_{\mathrm{ISM}}
    =\tfrac12\,\bm\vartheta^T\langle\bm\Xi^T\bm\Xi\rangle\bm\vartheta
     +\langle\nabla\cdot\bm\Xi\rangle^T\bm\vartheta,
    \qquad
    \bm\vartheta^\ast=-\langle\bm\Xi^T\bm\Xi\rangle^{+}\langle\nabla\cdot\bm\Xi\rangle,
    \label{eq:app_linear_ism}
\end{equation}
where \(\nabla\cdot\bm\Xi\) is the vector of column divergences, and the
superscript \(+\) denotes the Moore--Penrose pseudoinverse. The
minimization of Eq.~\eqref{eq:app_ism_objective} over all parameters starts from
\begin{equation}
    (\bm w,\kappa)=\bm\vartheta^\ast,
    \qquad
    \bm w_{\mathrm{out}}=0,
\end{equation}
so that the initial score is the best linear score \(\bm\Xi\bm\vartheta^\ast\), and
the hidden layers learn the corrections to it. The linear solve and the
training use the states saved every \(0.584\) time units, the coarsest interval of Fig.~\ref{fig:spin_km}, so that the same score
serves every sampling interval.
\end{revctext}

\subsection{\texorpdfstring{\revc{Conditional score}}{Conditional score}}
\label{app:conditional_score}

\begin{revctext}
With \(p(\bm x_0,\bm x_t)\) the joint density of lagged pairs,
\begin{equation}
    \bm s_{t|0}=\nabla_{\bm x_0}\log p(\bm x_0,\bm x_t)-\bm s(\bm x_0).
    \label{eq:app_conditional_score_def}
\end{equation}
Implicit score matching in \(\bm x_0\) for \(p(\bm x_0,\bm x_t)\), with the
stationary network score \(\widehat{\bm s}\) held fixed, gives
\begin{equation}
    \mathcal J_t(\theta)
    =\Bigl\langle
      \tfrac12\|\bm s^\theta_{t|0}\|^2
      +\nabla_{\bm x_0}\cdot\bm s^\theta_{t|0}
      +\widehat{\bm s}(\bm x_0)\cdot\bm s^\theta_{t|0}
     \Bigr\rangle_{\mathrm{obs}},
    \qquad
    \operatorname*{arg\,min}_{\bm s^\theta_{t|0}}\mathcal J_t
    =\nabla_{\bm x_0}\log p(\bm x_0,\bm x_t)-\widehat{\bm s}(\bm x_0),
    \label{eq:app_conditional_ism}
\end{equation}
where \(\langle\cdot\rangle_{\mathrm{obs}}\) averages over the observed pairs at
lag \(t\). The lags are weighted by \(t\), which compensates for the growth
\(\|\bm s_{t|0}\|^2\propto t^{-1}\) at short lags,
\begin{equation}
    \mathcal J(\theta)=\mathbb E_t\bigl[t\,\mathcal J_t(\theta)\bigr],
    \qquad
    t\ \text{uniform on the saved lags in}\ [t_{\min},t_{\max}].
\end{equation}
The model is a conservative field in \(\bm x_0\) of the local
form~\eqref{eq:app_local_energy}, without confining term,
\begin{equation}
    \bm s^\theta_{t|0}=\nabla_{\bm x_0}V_\theta,
    \qquad
    V_\theta=\sum_{i=1}^{N}u_\theta\bigl(\bm\xi_i,\bm\tau(t)\bigr),
    \qquad
    \bm\xi_i=\bm\xi\bigl(\bm x_{0,\mathcal N_i},\bm\delta_{\mathcal N_i}\bigr),
    \qquad
    \bm\delta=\frac{\bm x_t-\bm x_0}{\sqrt t},
    \label{eq:app_conditional_energy}
\end{equation}
where \(u_\theta\) is a network~\eqref{eq:app_residual_network} with input
\((\bm\xi_i,\bm\tau)\) and the lag enters through
\begin{equation}
    \bm\tau(t)=\bigl(v,\{\sin k\pi v,\cos k\pi v\}_{k=1}^{8}\bigr),
    \qquad
    v=\frac{\log(t/t_{\min})}{\log(t_{\max}/t_{\min})}.
\end{equation}
Derivatives act on \(\bm x_0\) at fixed \(\bm x_t\), so the
rules~\eqref{eq:app_forward_laplacian} start from
\begin{equation}
    \nabla_{\bm x_0}\bm\delta=-t^{-1/2}\,\bm I,
    \qquad
    \nabla_{\bm x_0}\bm\tau=0 .
\end{equation}

For the two-dimensional diffusion, \(L=3\), \(\dim\bm h^{(\ell)}=128\), and
\(0.01\leq t\leq1.5\), with
\begin{equation}
    \bm\xi=\bigl(\bm\zeta,\{\zeta_a\zeta_b\}_{a\leq b}\bigr),
    \qquad
    \bm\zeta=\bigl(\tilde{\bm x}_0,\tilde{\bm\delta}\bigr),
\end{equation}
where \(\tilde{\bm x}_0\) is standardized and \(\tilde{\bm\delta}\) is \(\bm\delta\)
divided by its root-mean-square value. For the soft-spin chain, \(r=2\), \(L=3\),
\(\dim\bm h^{(\ell)}=128\), and \(0.256\leq t\leq0.876\); the features are the
invariants~\eqref{eq:app_spin_invariants} of the ten vectors formed by the values
of \(\bm m_0\) and \(\bm\delta\) at the five sites of \(\mathcal N_i\), with the same
reflection average.
Neither score depends on the sampling interval, so the same networks are used
at every \(\Delta t\).

\paragraph*{\revc{Data-only check.}}
The conditional score is tested on the data alone through an identity satisfied
by the exact scores. For a smooth test function \(h(\bm x_0)\), an observable
\(\psi(\bm x_t)\), and \(\partial_i=\partial/\partial x_{0,i}\), the integral of a
total derivative vanishes,
\begin{equation}
    \int\!\!\int\psi(\bm x_t)\,
      \partial_i\bigl[h(\bm x_0)\,p(\bm x_0,\bm x_t)\bigr]\,
      \d\bm x_0\,\d\bm x_t=0,
\end{equation}
provided that \(h\,p\) vanishes on the boundary of the state space. Since
\(\partial_i p(\bm x_0,\bm x_t)=p(\bm x_0,\bm x_t)\,(\bm s+\bm s_{t|0})_i\) by
Eq.~\eqref{eq:app_conditional_score_def}, this becomes
\begin{equation}
    T_i[h,\psi](t)=
    \bigl\langle\psi(\bm x_t)\,
      \bigl[h(\bm x_0)\,(\bm s+\bm s_{t|0})_i+\partial_i h(\bm x_0)\bigr]
    \bigr\rangle_{\mathrm{obs}}=0 .
    \label{eq:app_conditional_ibp}
\end{equation}
We evaluate \(T_i\) on the observed pairs with the network scores
\(\widehat{\bm s}\) and \(\bm s^\theta_{t|0}\), and normalize it by its value for
\(\bm s^\theta_{t|0}=0\),
\begin{equation}
    R[h]=\left(
      \frac{\sum_{\psi,i,t}T_i[h,\psi](t)^2}
           {\sum_{\psi,i,t}T_i[h,\psi](t)^2\big|_{\bm s^\theta_{t|0}=0}}
    \right)^{1/2},
\end{equation}
where the sums run over the observables, the components, and the lags. For
exact scores, \(R[h]\) vanishes up to the sampling error of the averages in
Eq.~\eqref{eq:app_conditional_ibp}. The values are given in Eq.~\eqref{eq:app_cs_violation} for the two-dimensional diffusion and
in Appendix~\ref{app:soft_spin_details} for the soft-spin chain.
\end{revctext}

\section{Kramers--Moyal estimates}
\label{app:kramers_moyal}

\begin{revctext}
With \(\mathcal F\) and \(\mathcal D\) the linear spans of the vector fields of
the drift library and of the symmetric matrix fields of the diffusion library,
and with the increments \(\Delta\bm x=\bm x(t+\Delta t)-\bm x(t)\) over the
sampling interval, the Kramers--Moyal estimates are
\begin{equation}
\begin{aligned}
    \widehat{\bm F}&=\operatorname*{arg\,min}_{\bm F\in\mathcal F}
    \left\langle\left|\frac{\Delta\bm x}{\Delta t}-\bm F(\bm x)\right|^2\right\rangle,\\
    \widehat{\bm D}&=\operatorname*{arg\,min}_{\bm D\in\mathcal D}
    \left\langle\left\|\frac{\delta\bm x\,\delta\bm x^T}{2\Delta t}
    -\bm D(\bm x)\right\|_F^2\right\rangle,
\end{aligned}
\label{eq:app_km_fits}
\end{equation}
where \(\bm x=\bm x(t)\) and the averages are taken over the sampled pairs of
states. The conventional estimate uses \(\delta\bm x=\Delta\bm x\), and the
drift-centered estimate uses
\(\delta\bm x=\Delta\bm x-\widehat{\bm F}(\bm x)\Delta t\)~\cite{Gottschall2008}.
The models
\(\d\bm x=\widehat{\bm F}\,\d t+\sqrt{2\widehat{\bm D}}\,\d\bm W\) are simulated
with the protocols of the score-based models, and because \(\widehat{\bm D}\) is
not constrained to be positive definite, the noise amplitude
\(\sqrt{2\widehat{\bm D}}\) is regularized. In
the two-dimensional example, \(\widehat{\bm D}\) is replaced by
\(\widehat{\bm D}+(10^{-6}-\lambda_{\min})\bm I\) wherever its smallest
eigenvalue \(\lambda_{\min}\) falls below \(10^{-6}\). In the soft-spin chain,
the noise amplitude of each site is \(\sqrt2\,\bm L\), where \(\bm L\) is the
lower-triangular factor of the onsite block of \(\widehat{\bm D}\) given by the
Cholesky recursion with negative squared diagonal entries set to zero,
\begin{equation}
    L_{kk}=\Bigl[\max\Bigl(\widehat D_{kk}-\sum_{j<k}L_{kj}^2,\,0\Bigr)\Bigr]^{1/2},
    \qquad
    L_{ik}=\frac{\widehat D_{ik}-\sum_{j<k}L_{ij}L_{kj}}{L_{kk}},
    \quad i>k,
\end{equation}
and \(L_{ik}=0\) for \(i>k\) whenever \(L_{kk}=0\). For a positive-definite
\(\widehat{\bm D}\), this gives \(\bm L\bm L^T=\widehat{\bm D}\).

In the two-dimensional example, the conventional estimate has smaller
diffusion errors and, up to \(\Delta t=0.2\), smaller target-correlation errors,
but larger stationary-density errors. In the soft-spin chain, its
transverse-diffusion errors are at least \(56\%\) at every interval. Its
correlations at \(\Delta t=0.0365\) are closer to the data than those of the
drift-centered estimate but still exceed the sampling level by an order of
magnitude, and its simulations diverge from \(\Delta t=0.073\).
\end{revctext}

\section{Technical details for the examples}
\label{app:examples_details}

\subsection{\texorpdfstring{\rev{Cox--Ingersoll--Ross diffusion}}{Cox--Ingersoll--Ross diffusion}}
\label{app:cir_benchmark_details}

\begin{revtext}
\begin{revbtext}
Write \(p=p_{\mathrm{ss}}\), \(\nu=\kappa\mu/\gamma>2\),
\(\beta=\kappa/\gamma\), and \(z=e^{-\kappa t}\); brackets denote stationary
averages.

\paragraph*{Stationary quantities and correlations.}
Zero stationary flux gives
\begin{equation}
\begin{aligned}
    0&=\kappa(\mu-x)p-(\gamma xp)',
    \\
    s&=\frac{p'}p=\frac{\nu-1}{x}-\beta,
    \qquad p(x)=\frac{\beta^\nu}{\Gamma(\nu)}x^{\nu-1}e^{-\beta x},
    \\
    \mu_a&=\int_0^\infty x^ap(x)\,\d x
    =\frac{\Gamma(\nu+a)}{\Gamma(\nu)\beta^a},
    \quad a>-\nu,
    \qquad V=\frac{\nu}{\beta^2}.
\end{aligned}
    \label{eq:app_cir_zero_flux}
\end{equation}
The zero flux also implies reversibility. The conditional mean obeys
\begin{equation*}
\begin{aligned}
    m_t(x)&=\mathbb E[X_t\mid X_0=x],
    \qquad \partial_t m_t=\kappa(\mu-m_t),\quad m_0(x)=x,
    \\
    m_t(x)&=\mu+z(x-\mu).
\end{aligned}
\end{equation*}
Hence
\begin{equation}
\begin{aligned}
    C_{f,x}(t)&=C_{x,f}(t)=\langle f m_t\rangle
    =\mu\langle f\rangle+z\langle(x-\mu)f\rangle,
    \\
    \dot C_{f,x}(t)&=-\kappa z\langle(x-\mu)f\rangle,
    \\
    \Phi&=-\dot C_{x,x}(0^+)=\kappa V=\gamma\mu.
\end{aligned}
    \label{eq:app_cir_correlations}
\end{equation}

\paragraph*{Conjugate observables.}
For smooth, linearly independent, centered fields \(H_k\) of polynomial growth
with \(q_k\in L^2(p)\), integration by parts gives
\begin{equation*}
\begin{aligned}
    q_k&=H_ks+H_k'=\frac{(pH_k)'}p,
    \qquad \langle fq_k\rangle=-\langle f'H_k\rangle,
    \\
    \langle q_k\rangle&=\langle xq_k\rangle=0,
    \qquad C_{q_k,x}(t)=\dot C_{q_k,x}(t)=0.
\end{aligned}
\end{equation*}
For the monomials of Sec.~\ref{subsec:cir_benchmark},
\(q_k\sim-\mu_k(\nu-1)/x\) near zero, so \(q_k^2p=O(x^{\nu-3})\) is
integrable for \(\nu>2\). With \(H_{\bm a}=\sum_k a_kH_k\) and
\(q_{\bm a}=\sum_k a_kq_k\),
\begin{equation}
    \bm a^T\bm G_x\bm a
    =\langle q_{\bm a}^2\rangle
    =\left\langle\left(\frac{(pH_{\bm a})'}p\right)^2\right\rangle.
    \label{eq:app_cir_gram}
\end{equation}
This vanishes only if \(pH_{\bm a}\) is constant. Its decay at infinity forces
\(H_{\bm a}=0\), and linear independence gives \(\bm a=0\); thus
\(\bm G_x\succ0\).

\paragraph*{Invertibility at finite lags.}
Reversibility makes the strongly continuous semigroup
\(K_t h(x)=\mathbb E[h(X_t)\mid X_0=x]\) self-adjoint on \(L^2(p)\), so
\begin{equation}
    \bm a^T\bm B(t)\bm a
    =\langle q_{\bm a}K_tq_{\bm a}\rangle
    =\|K_{t/2}q_{\bm a}\|_{L^2(p)}^2.
    \label{eq:app_cir_finite_lag_positivity}
\end{equation}
For finite \(t>0\),
\begin{equation*}
    K_t h=0
    \ \Longrightarrow\ 
    \|K_{t/2}h\|_{L^2(p)}^2=\langle hK_th\rangle=0
    \ \Longrightarrow\ 
    K_{t/2^n}h=0\quad(n\ge1).
\end{equation*}
Strong continuity at zero implies \(h=0\). Thus \(K_t\) is injective, and
\(q_{\bm a}\neq0\) for \(\bm a\neq0\) makes
Eq.~\eqref{eq:app_cir_finite_lag_positivity} strictly positive:
\(\bm B(t)\succ0\) for every finite \(t>0\), with \(\bm B(0)=\bm G_x\).

\paragraph*{Exact recovery.}
If \(x-\mu=\sum_k b_kH_k\), then
\begin{equation*}
\begin{aligned}
    \sum_k b_kq_k&=(x-\mu)s+1=\nu-\beta x-\mu s,
    \\
    C_{q_j,s}(t)&=-\frac1\mu\sum_k b_kB_{jk}(t),
    \\
    y_j(t)&=\dot C_{q_j,x}(t)-\Phi C_{q_j,s}(t)
    =\frac{\Phi}{\mu}\sum_k B_{jk}(t)b_k.
\end{aligned}
\end{equation*}
Consequently,
\begin{equation}
    \bm y(t)=\frac{\Phi}{\mu}\bm B(t)\bm b,
    \qquad \bm c=\frac{\Phi}{\mu}\bm b=\gamma\bm b,
    \qquad M_{\bm c}=\Phi+\gamma(x-\mu)=\gamma x.
    \label{eq:app_cir_general_basis}
\end{equation}
For the monomial basis, \(\bm b=\bm e_1\), recovering
Eq.~\eqref{eq:cir_fixed_basis_target}.
\end{revbtext}
\end{revtext}

\subsection{Technical details for the two-dimensional affine multiplicative-noise benchmark}
\label{app:affine_multiplicative_2d_details}

\begin{revtext}
\begin{revbtext}
\paragraph*{Data and score estimates.}
The noise matrices in Eq.~\eqref{eq:affine_2d_reference_sde_main} are
\begin{equation}
\begin{aligned}
    \bm\Lambda_0&=\begin{pmatrix}0.60&0.10\\0.08&0.55\end{pmatrix},
    &\bm\Lambda_1&=\begin{pmatrix}0.36&0.12\\0.15&-0.06\end{pmatrix},\\
    \bm\Lambda_2&=\begin{pmatrix}-0.09&0.21\\0.18&0.30\end{pmatrix}.
\end{aligned}
\label{eq:affine_2d_B_coefficients_main}
\end{equation}
These matrices have generic entries, with no symmetry, and their linear
terms are comparable to the constant one. The diffusion tensor therefore varies
strongly over the region visited by the data, while the affine form keeps it
positive semidefinite and quadratic in the coordinates. The rotation rate
\(\omega=1.1\) is comparable to the relaxation rate of the gradient drift near
the origin, which is one, so that both parts of the drift shape the dynamics.
The data comprise \(36\) Euler--Maruyama trajectories of duration \(3000\),
with step \(10^{-3}\) and saving interval \(\Delta t=10^{-2}\). Discarding
the first \(300\) time units of each leaves \(T_{\mathrm{data}}=97200\).
\revc{The stationary network score \(\widehat{\bm s}\) gives the Stein check}
\begin{equation}
\left\langle(\bm x-\langle\bm x\rangle)\widehat{\bm s}^{T}\right\rangle
\simeq-\operatorname{diag}(\revc{0.998},\revc{0.996}),
\end{equation}
\revc{with off-diagonal entries below \(0.003\).}
For the fixed-basis calibration, correlation derivatives are estimated by
local cubic fits with bandwidth \(3\Delta t\)
(Appendix~\ref{app:local_polynomial_cdot_estimation}). \revc{The neural
calibration uses cubic smoothing splines in the lag; among the smoothing
levels whose fit residual exceeds the smallest by at most \(2\%\), it takes the
one with the smoothest derivative.} Applying the \revc{local cubic}
estimator at zero lag to the coordinate correlations, with a one-sided
fitting window, gives the common mean mobility
\begin{equation}
    \widehat{\bm\Phi}
    =\begin{pmatrix}0.197&0.186\\-0.076&0.160\end{pmatrix},
\label{eq:affine_2d_phi_main}
\end{equation}
with jackknife errors below \(10^{-3}\).

\paragraph*{Reference fields and field errors.}
The reference diffusion is \(\bm D_{\mathrm{ref}}=\bm\Lambda\bm\Lambda^T/2\).
With \(\bm F=-\nabla U+\omega\bm J\bm x\), the reference circulation is
reconstructed by finite-difference least squares on a grid from
\begin{equation}
    \bm J\bigl(\nabla r_{\mathrm{ref}}+r_{\mathrm{ref}}\widehat{\bm s}\bigr)
    =\bm F-\bm D_{\mathrm{ref}}\widehat{\bm s}
          -\nabla\cdot\bm D_{\mathrm{ref}}.
\label{eq:app_affine_circulation}
\end{equation}
The reconstructed reference inherits the estimation error of the stationary
score. For the forward comparisons, the circulation is reconstructed using
the same \revc{network} score as the calibrated models. The empirical stationary
averages defining the relative field errors are
\begin{equation}
\begin{aligned}
    e_D^2&=\frac{\langle\|\bm D-\bm D_{\mathrm{ref}}\|_F^2\rangle}
    {\langle\|\bm D_{\mathrm{ref}}-\langle\bm D_{\mathrm{ref}}\rangle\|_F^2\rangle},\\
    e_r^2&=\frac{\langle(r-r_{\mathrm{ref}})^2\rangle}
    {\langle(r_{\mathrm{ref}}-\langle r_{\mathrm{ref}}\rangle)^2\rangle}.
\end{aligned}
\label{eq:app_affine_field_errors}
\end{equation}
The best-approximation curve in Fig.~\ref{fig:affine_selection}(c) is obtained
by least-squares regression of
\(r_{\mathrm{ref}}-\langle r_{\mathrm{ref}}\rangle\) on the centered
monomials of degree at most \(P_R\), using the same stationary averages.
\revc{The drift error \(e_F\) of Fig.~\ref{fig:affine_km} is defined in the same
way from the exact drift \(\bm F\) and the drift of a model, which is
\(\bm M\widehat{\bm s}+\nabla\cdot\bm M\) for the score-based models and
\(\widehat{\bm F}\) for the Kramers--Moyal models.}

\paragraph*{Fixed-basis calibration.}
The centered basis fields are
\begin{equation}
\begin{aligned}
    \bm H_k&=h_{a_kb_k}\bm T_{\ell_k},
    &h_{ab}&=u^av^b-\langle u^av^b\rangle,\\
    \bm T_1&=\bm e_1\bm e_1^T,
    &\bm T_2&=\bm e_2\bm e_2^T,\\
    \bm T_3&=\bm e_1\bm e_2^T+\bm e_2\bm e_1^T,
    &\bm T_4&=\bm J,
\end{aligned}
\label{eq:app_affine_basis}
\end{equation}
where \(\bm e_i\) are coordinate unit vectors. The symmetric fields use
\(1\leq a+b\leq P_D\), and the circulation fields use
\(1\leq a+b\leq P_R\), giving
\begin{equation}
    \bm M_{\bm c}=\widehat{\bm\Phi}+\sum_{k=1}^{K}c_k\bm H_k,
    \qquad K=\tfrac32P_D(P_D+3)+\tfrac12P_R(P_R+3).
\label{eq:affine_2d_K}
\end{equation}
For coordinate initial observables, the conjugate observables form the vector
\(\bm T_{\ell_k}^{T}(\nabla h_{a_kb_k}+h_{a_kb_k}\widehat{\bm s})\);
for quadratics they follow from Eq.~\eqref{eq:basis_quadratic_response}.
Each basis field is rescaled by the root-mean-square norm of its coordinate
conjugate observables before forming the matrices.

The nine target monomials and the conjugate observables of all \(47\) fields
in the largest candidate basis, \((P_D,P_R)=(3,5)\), give the common equations
used to compare candidates:
\begin{equation}
\begin{aligned}
 B^{\mathrm{tar}}_{\ell mn,k}
    &=\langle\psi_m(\bm x_{t_\ell})q_{kn}(\bm x_0)\rangle,\\
 y^{\mathrm{tar}}_{\ell mn}
    &=\dot C_{\psi_m,\phi_n}(t_\ell)
      -C_{\psi_m,\mathcal L_{\widehat{\bm\Phi}}^\dagger\phi_n}(t_\ell),\\
 B^{q}_{\ell j,k}
    &=\sum_n\langle q_{jn}(\bm x_{t_\ell})q_{kn}(\bm x_0)\rangle,\\
 y^{q}_{\ell j}
    &=\sum_n\left[\dot C_{q_{jn},\phi_n}(t_\ell)
      -C_{q_{jn},\mathcal L_{\widehat{\bm\Phi}}^\dagger\phi_n}(t_\ell)\right].
\end{aligned}
\label{eq:app_affine_calibration_blocks}
\end{equation}
Here \(j=1,\ldots,47\), while \(k\) runs over the candidate basis. The
twenty fitted lags give \(1300\) equations with coordinate initial observables
and \(1840\) after adding \(u^2,uv,v^2\). Within each block \(b\), the
diagonal weights are
\begin{equation}
    W_{\alpha\alpha}
    =\left[\max\!\left(v_\alpha,
             0.01\,\operatorname{median}_{\beta\in b}v_\beta\right)\right]^{-1},
    \qquad v_\alpha=\widehat{\operatorname{Var}}(\widehat y_\alpha).
\end{equation}
Variances and matrix diagnostics are estimated over trajectories as in
Appendix~\ref{app:fixed_basis_details}.
Candidate bases are compared in six folds of six trajectories, fitting and
evaluating the common conjugate-observable equations on separate groups. Among
identifiable candidates, we retain the smallest whose mean residual differs
from the minimum by no more than one standard error of that difference across
folds. The selected \((2,2)\) model is refitted to both blocks using all data.
Its unconstrained diffusion has minimum eigenvalue \revc{\(0.016\)} on the sampled
region, so the positivity constraint is inactive.
\end{revbtext}
\end{revtext}

\begin{revtext}
\begin{revbtext}
\paragraph*{Neural calibration and conditional-score accuracy.}
The mobility has two hidden layers of width \(128\), with outputs for the three
entries of \(\bm L_\theta\) and the scalar \(r_\theta\) in
Eq.~\eqref{eq:neural_M_DR_parametrization}; \(\varepsilon=10^{-5}\).
Training uses \(250\) Adam epochs, learning rate \(3\times10^{-4}\),
\(32768\) pairs per lag, batches of four lags, and \(32768\) states for the
mean penalty. At the thirty fitted lags, with \(t_{\min}=0.03\), the weights in
Eq.~\eqref{eq:deltaM_neural_loss} are
\begin{equation}
\begin{aligned}
    S_{mn}&=\max\!\left[
       \left(\frac1{30}\sum_\ell\widehat y_{mn}(t_\ell)^2\right)^{1/2},10^{-4}
       \right],\\
    w_{mn,t}&=\frac{w_t}{S_{mn}^2},\qquad
    w_t=\frac{(t_{\min}/t)^{1/2}}
                {\frac1{30}\sum_\ell(t_{\min}/t_\ell)^{1/2}}.
\end{aligned}
\end{equation}
The mean penalty is evaluated entrywise as
\begin{equation}
    \mathcal J_{\mathrm{mean}}
    =\frac14\sum_{i,j=1}^{2}
    \frac{\langle\delta M_{\theta,ij}\rangle^2}
         {\max(|\widehat\Phi_{ij}|,10^{-4})^2},
    \qquad \lambda_{\mathrm{mean}}=0.1,
\end{equation}
and the parameter penalty is \(10^{-6}\) times the mean of the squared network
parameters.

The data-only check of Appendix~\ref{app:conditional_score}, with
\(h\in\{1,x,y\}\), the nine target future observables as \(\psi\),
\(8\times10^5\) pairs per lag, and
\(t\in\{0.03,0.05,0.1,0.2,0.3,0.5,0.75,1,1.5\}\), gives
\begin{equation}
 \bigl(R[1],R[x],R[y]\bigr)=(\revc{0.020},\revc{0.090},\revc{0.068}).
\label{eq:app_cs_violation}
\end{equation}

\paragraph*{Forward validation and sampling uncertainty.}
Each model is integrated with \(1024\) Euler--Maruyama trajectories, step
\(0.005\), duration \(110\), and initial discard \(10\), giving
\(T_{\mathrm{model}}=102400\). Initial states and noise are shared across
models. Basis divergences are analytic; the neural mobility and its divergence
are tabulated on a \(601\times601\) grid spanning the observed region plus
\(10\%\), with centered differences for the divergence.

The stationary mean of each observable is estimated separately from the
observed records and each model simulation. Correlations are computed from
fluctuations about these means, using the same observed-variance normalization,
\begin{equation}
    \rho_{mn}^{a}(t)=
    \frac{\operatorname{Cov}_{a}(\psi_m(\bm x_t),\phi_n(\bm x_0))}
      {\sigma_{\psi_m,\mathrm{data}}\sigma_{\phi_n,\mathrm{data}}},
    \qquad a\in\{\mathrm{data},\mathrm{model}\}.
\end{equation}
The RMSE for each pair and its mean over an observable group
\(\mathcal G\) are
\begin{equation}
\begin{aligned}
    E_{mn}&=\left[\frac1{N_t}\sum_{0\leq t_\ell\leq4}
       \bigl(\rho_{mn}^{\mathrm{model}}(t_\ell)
             -\rho_{mn}^{\mathrm{data}}(t_\ell)\bigr)^2\right]^{1/2},\\
    E_{\mathcal G}&=\frac1{|\mathcal G|}\sum_{(m,n)\in\mathcal G}E_{mn}.
\end{aligned}
\label{eq:app_affine_correlation_rmse}
\end{equation}
Here \(N_t=401\) covers the full displayed lag range at spacing \(0.01\).
The observed standard deviations are held fixed across models, so the error
also retains differences in their stationary variances. The groups in
Table~\ref{tab:affine_forward} are the eighteen targets, the ten quartic
correlations with the coordinates, and the squared-amplitude autocorrelation.
At each lag, standard errors in Fig.~\ref{fig:affine_forward_cphi} are
estimated from the variation across trajectories. To estimate the discrepancy
due to finite sampling, we compare correlations computed from half of the
observed record with those computed from the full record, using the same error
measure defined above. Denoting this discrepancy by \(E_{1/2}\), and assuming
that sampling variances are inversely proportional to record duration, we
estimate the sampling floor as
\begin{equation}
    E_{\mathrm{floor}}
      =E_{1/2}\left(1+\frac{T_{\mathrm{data}}}{T_{\mathrm{model}}}\right)^{1/2},
\label{eq:app_affine_floor}
\end{equation}
where \(T_{\mathrm{data}}\) and \(T_{\mathrm{model}}\) are the total retained
durations of the observed and simulated trajectories, respectively. Sampling
uncertainty is reported separately from RMSE through the pointwise bands in
Fig.~\ref{fig:affine_forward_cphi}.
Stationary-density errors are measured by the \(L^1\) distance between
normalized \(60\times60\) histograms.
\end{revbtext}
\end{revtext}

\begin{table*}[t]
    \centering
    \small
    \caption{\revb{Errors for the models compared in
    Sec.~\ref{subsec:affine_multiplicative_2d_results}, using the same
    \revc{network} stationary score: mean channelwise root-mean-square correlation
    errors on observed-variance scales over \(0\leq t\leq4\),
    Eq.~\eqref{eq:app_affine_correlation_rmse}, and stationary-density \(L^1\) errors.
    The fixed-basis model uses \((P_D,P_R)=(2,2)\).}}
    \label{tab:affine_forward}
    \begin{tabular}{lcccc}
        \hline\hline
        model & targets & quartic & \(\varrho\) ACF & density \(L^1\)\\
        \hline
        \(\bm M=\widehat{\bm\Phi}\) & \revc{0.0294} & \revc{0.0325} & \revc{0.0693} & \revc{0.014}\\
        basis, selected & \revc{0.0028} & \revc{0.0022} & \revc{0.0028} & \revc{0.014}\\
        network, conditional score & \revc{0.0034} & \revc{0.0028} & \revc{0.0031} & \revc{0.013}\\
        \(\bm M_{\mathrm{ref}}\) & \revc{0.0029} & \revc{0.0022} & \revc{0.0029} & \revc{0.014}\\
        \hline\hline
    \end{tabular}
\end{table*}

\begin{figure*}[p]
    \centering
    \includegraphics[width=\textwidth]{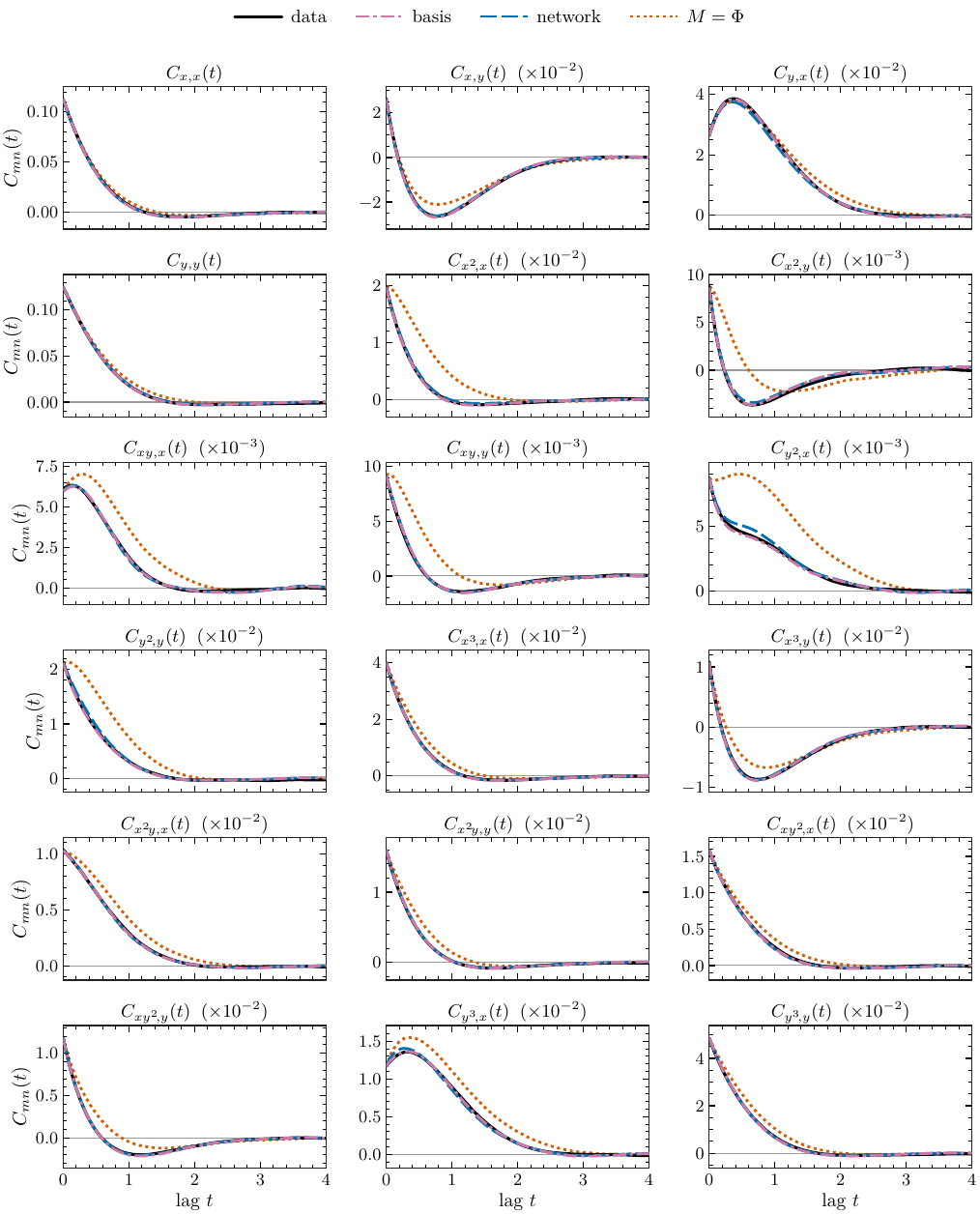}
    \caption{\revb{All eighteen target correlations of the two-dimensional
    diffusion, with the models and simulation protocol of
    Fig.~\ref{fig:affine_forward_cphi}. Correlations are computed from
    fluctuations about the stationary means of the observables and rescaled
    by the powers of ten in their titles.}}
    \label{fig:affine_forward_all}
\end{figure*}

\clearpage

\subsection{Technical details for the periodic soft-spin benchmark}
\label{app:soft_spin_details}

\begin{revtext}
\begin{revbtext}
\paragraph*{System and data.}
The parameters of Eq.~\eqref{eq:soft_spin_ll_sde_results} are
\begin{equation}
\begin{gathered}
 (N,\lambda,m_\ast,J,K_z,\Theta)=(12,10,1,0.35,0.5,0.2),\\
 (\gamma,\alpha_\perp,\alpha_\parallel,\varepsilon)
 =(4,0.25,0.03,10^{-2}).
\end{gathered}
\end{equation}
With \(\bm u_i=\bm m_{i-1}+\bm m_{i+1}-2\bm m_i\), the Gibbs score is
\begin{equation}
\begin{aligned}
 \bm s_i&=\beta_1\bm m_i+\beta_3\varrho_i\bm m_i
             +\beta_J\bm u_i+\beta_zm_{i,z}\bm e_z,\\
 (\beta_1,\beta_3,\beta_J,\beta_z)&=(50,-50,1.75,2.5).
\end{aligned}
\label{eq:soft_spin_score_results}
\end{equation}
Define
\(\mathsf T_i=\varrho_i\bm I-\bm m_i\bm m_i^T\),
\(\mathsf P_i=\bm m_i\bm m_i^T\), and
\(\mathsf C_i=[\bm m_i]_\times\). The true onsite mobility is
\begin{equation}
\begin{aligned}
 \bm M_{ii}&=c_0\bm I+c_\perp\mathsf T_i
              +c_\parallel\mathsf P_i+c_\times\mathsf C_i,\\
 (c_0,c_\perp,c_\parallel,c_\times)&=(0.002,0.05,0.006,-0.8).
\end{aligned}
\end{equation}
The data are \(72\) independent Euler--Maruyama trajectories with step
\revc{\(2.5\times10^{-5}\)}, saving interval \(\Delta t=0.0365\), and \(277778\)
saved states per trajectory. Discarding the first \(10\%\) leaves approximately
\(1.8\times10^7\) configurations and total duration
\(T_{\mathrm{data}}\simeq6.6\times10^5\). The state ordering is
\((m_{1x},m_{1y},m_{1z},\ldots,m_{Nz})\). The site-averaged magnetization
correlations first fall below \(0.05\) at lags \revc{\(2.85\), \(2.88\), and
\(316\)} for \(x,y,z\), respectively. \revc{The small step keeps the bias of the
Euler--Maruyama stationary density, which grows linearly with the step and is
amplified by the precession, below the sampling uncertainty.}

\paragraph*{Stationary score.}
\revc{With respect to the exact Gibbs score~\eqref{eq:soft_spin_score_results},
the network score \(\widehat{\bm s}\) has the relative error
\begin{equation}
    \frac{\langle\|\widehat{\bm s}-\bm s\|^2\rangle^{1/2}}{\langle\|\bm s\|^2\rangle^{1/2}}
    =1.1\times10^{-3},
\end{equation}
and it satisfies \(\langle\widehat s_{i,c}\,m_{i,c}\rangle=-1\) to within about one
standard error.} For the functions \(\varphi\in\{m_{i,x}^2,m_{i,y}^2,m_{i,z}^2,\varrho_i\}\),
the relative Stein discrepancies, averaged over sites, are
\begin{equation}
\begin{aligned}
 \delta_\varphi
 &=1+\frac{\langle\widehat{\bm s}\cdot\nabla\varphi\rangle}
                 {\langle\Delta\varphi\rangle},\\
 (\delta_{m_x^2},\delta_{m_y^2},\delta_{m_z^2},\delta_\varrho)
 &\simeq(\revc{0.0007},\revc{0.0019},\revc{0.0030},\revc{0.0018}).
\end{aligned}
\end{equation}
An exact stationary score gives \(\delta_\varphi=0\)\revc{; the jackknife standard
errors are \(0.0013\) to \(0.0035\)}.

\paragraph*{Fixed-basis equations and selection.}
The uncentered expansion is \(\bm M_{ii}=\sum_{k=1}^{19}c_k\bm F_k(\bm m_i)\).
Its fields are all products \(p\bm G\) of total degree at most two, with
\begin{equation}
\begin{aligned}
 p&\in\{1,m_z,\varrho,m_z^2\},\\
 \bm G&\in\left\{\bm I,\bm e_z\bm e_z^T,[\bm e_z]_\times,
 \tfrac12(\bm m\bm e_z^T+\bm e_z\bm m^T),
 [\bm m]_\times,[\bm m\times\bm e_z]_\times,\bm m\bm m^T\right\}.
\end{aligned}
\end{equation}
For coordinate initial observables, the conjugate vector associated with
\(\bm F_k\) is
\begin{equation}
 \bm v_{k,i}=\bm F_k^T\widehat{\bm s}_i
                 +\nabla_{\bm m_i}\cdot\bm F_k^T.
\label{eq:app_spin_responses}
\end{equation}
The coordinate Gram matrix has smallest normalized eigenvalue
\revc{\(7.2\times10^{-4}\)} and \(\eta_G=\revc{0.017}\) with the network score.
Thus no additional initial observables are required.
Writing \(\widetilde f=f-\langle f\rangle\), the target-coordinate equations are
\begin{equation}
\begin{aligned}
 B_{(\ell,d,r),k}
 &=\frac1N\sum_i\left\langle
 \widetilde m_{i,d}(t_\ell)(\bm v_{k,i+r})_d(0)\right\rangle,\\
 y_{\ell,d,r}
 &=\left.\frac{\d}{\d t}\frac1N\sum_i\left\langle
 \widetilde m_{i,d}(t)\widetilde m_{i+r,d}(0)\right\rangle\right|_{t=t_\ell},
 \qquad r=0,\pm1.
\end{aligned}
\label{eq:app_spin_design_targets}
\end{equation}
The conjugate-observable equations sum over coordinate components:
\begin{equation}
\begin{aligned}
 B_{(\ell,j),k}
 &=\frac1N\sum_i\left\langle
 \widetilde{\bm v}_{j,i}(t_\ell)\cdot\bm v_{k,i}(0)\right\rangle,\\
 y_{\ell,j}
 &=\left.\frac{\d}{\d t}\frac1N\sum_i\left\langle
 \widetilde{\bm v}_{j,i}(t)\cdot\widetilde{\bm m}_i(0)
 \right\rangle\right|_{t=t_\ell}.
\end{aligned}
\label{eq:app_spin_design_responses}
\end{equation}
The future-observable libraries remain unchanged as mobility fields are
removed; in particular, the conjugate library retains the observables
generated by all nineteen fields. Calibration uses the twenty-four lags
\(t_\ell=\ell\Delta t\), with \(\Delta t=0.0365\). Correlation derivatives
are estimated by local cubic fits with bandwidth \(h=\Delta t\)
(Appendix~\ref{app:local_polynomial_cdot_estimation}). Equations are weighted
by the inverse jackknife variances of these derivatives. Within each
observable library, variances below \(1\%\) of the median are replaced by
that threshold to prevent excessively large weights.

At each reduction step, we remove the field with the smallest ratio of
coefficient magnitude to jackknife standard error. Candidate bases are
compared using six groups of twelve trajectories, evaluating the calibration
equations on trajectories excluded from each fit. We retain the smallest
candidate whose mean error exceeds the minimum by no more than one standard
error of the difference. For the reported model, marked by black stars in
Fig.~\ref{fig:spin_accuracy}, this procedure selects \revc{four} fields using the
conjugate-observable equations\revc{, those of the true mobility}; their
coefficients are then refitted using both the conjugate-observable and target
equations.

\revc{During selection, the diffusion tensor is left unconstrained, so that
subsets with indefinite diffusion tensors can be identified}
[Fig.~\ref{fig:spin_calibration}(b)]. \revc{For the reported model, the
positivity constraint of Eq.~\eqref{eq:basis_wls} is inactive: the unconstrained
minimum eigenvalue over the sampled region is \(0.0049\).} The
corresponding coefficient-identification diagnostic is approximately
\(\eta_B=\revc{0.16}\) with the targets alone and \revc{\(0.012\)} with the conjugate
observables.

\paragraph*{Constant-mobility estimate.}
The fixed-basis fit above determines the full mobility without using
\(\bm\Phi\). For the constant-mobility comparison and the correction used
in neural calibration, we use a common estimate of \(\bm\Phi\), obtained
from the initial coordinate-correlation slopes by one-sided local cubic fits
(Appendix~\ref{app:local_polynomial_cdot_estimation}). The slopes are
averaged over sites and constrained to the form required by axial symmetry,
\begin{equation}
 \bm\Phi_{ij}=\delta_{ij}
 \begin{pmatrix}a&-b&0\\b&a&0\\0&0&c\end{pmatrix}.
\label{eq:soft_spin_axial_class}
\end{equation}
Their bandwidth dependence [Fig.~\ref{fig:spin_calibration}(a)] is
\begin{equation}
\begin{array}{c|rrrrr|r}
 h/\Delta t&0.75&1&1.5&2&3&\text{true mean}\\ \hline
 a&\revc{0.04152}&\revc{0.04149}&\revc{0.04114}&\revc{0.04041}&\revc{0.03834}&\revc{0.04195}\\
 c&\revc{0.02964}&\revc{0.02962}&\revc{0.02949}&\revc{0.02933}&\revc{0.02887}&\revc{0.02953}
\end{array}
\label{eq:app_spin_phi_bandwidth}
\end{equation}
At the chosen bandwidth \(h=\Delta t\), \(a=\revc{0.04149(7)}\),
\(b=\revc{-0.009(6)}\), and \(c=\revc{0.02962(15)}\), where parentheses give jackknife
standard errors. The slope blocks coupling different sites are below
\revc{\(1.5\%\)} of the onsite block. The common estimate is therefore
\begin{equation}
 \bm\Phi_{ij}=\delta_{ij}
 \begin{pmatrix}
 \revc{0.04149}&\revc{0.009}&0\\\revc{-0.009}&\revc{0.04149}&0\\0&0&\revc{0.02962}
 \end{pmatrix}.
\label{eq:app_soft_spin_phi_final_block}
\end{equation}
This matrix defines both the constant closure \(\bm M=\bm\Phi\) and the
correction \(\delta\bm M_\theta=\bm M_\theta-\bm\Phi\) in the
conditional-score constraints.

\paragraph*{Neural calibration.}
The initial observables are the spin coordinates. With
\(\bm a_i=m_{i,z}\bm e_z\), the future library comprises
\begin{equation}
\begin{aligned}
 \{\bm\psi_i^{(a)}\}_{a=1}^{11}=\bigl\{&
 b_0(\varrho_i)\bm m_i,b_1(\varrho_i)\bm m_i,b_2(\varrho_i)\bm m_i,\\
 &\bm u_i,\mathsf T_i\bm u_i,\mathsf P_i\bm u_i,\mathsf C_i\bm u_i,\\
 &\bm a_i,\mathsf T_i\bm a_i,\mathsf P_i\bm a_i,\mathsf C_i\bm a_i\bigr\},
\end{aligned}
\label{eq:soft_spin_eleven_family_library}
\end{equation}
where \(b_0,b_1,b_2\) are obtained by orthonormalizing
\(1,\varrho,\varrho^2\) under the empirical stationary distribution.
The first family contains the target coordinates. With the exact stationary
score in Eq.~\eqref{eq:soft_spin_score_results}, these families also span the
conjugate observables associated with the four true mobility fields.

Averaging over sites gives the correlations
\begin{equation}
\begin{aligned}
 C_{ac;dr}(t)&=\frac1N\sum_i
 \left\langle\widetilde\psi_{i,c}^{(a)}(t)
 \widetilde m_{i+r,d}(0)\right\rangle,\\
 &a=1,\ldots,11,\qquad c,d\in\{x,y,z\},\qquad r=0,\ldots,11,
\end{aligned}
\label{eq:app_soft_spin_eleven_family_correlation}
\end{equation}
where tildes denote subtraction of stationary means. These provide \(1188\)
scalar correlation curves. Their derivatives are estimated by local cubic
fits with bandwidth \(3\Delta t=0.1095\), \revc{using every second saved state
as the initial time of the pairs, about \(9\times10^{6}\) initial configurations,}
at the support lags \(t_\ell=\ell\Delta t\), \(\ell=0,\ldots,\revc{36}\). Calibration
uses the eighteen lags \(\ell=7,\ldots,24\), corresponding to
\(0.2555\leq t\leq0.876\).

The mobility network has four layers of width \(256\) with swish activations
and is shared by all sites. Its input is
\begin{equation}
 \xi_i=(\bm z_{i-1},\bm z_i,\bm z_{i+1},\|\bm z_i\|^2),
\end{equation}
where \(\bm z_i\) denotes the spin \(\bm m_i\) with each component
standardized by its stationary mean and standard deviation, pooled over sites.
Its nine outputs determine the onsite block through
\begin{equation}
\begin{aligned}
 \bm L_i&=0.40
 \begin{pmatrix}
 \operatorname{softplus}(y_1)&0&0\\
 y_2&\operatorname{softplus}(y_3)&0\\
 y_4&y_5&\operatorname{softplus}(y_6)
 \end{pmatrix}+10^{-5}\bm I,\\
 \bm k_i&=7.60(y_7,y_8,y_9)^T,\\
 (\bm M_\theta)_{ij}&=\delta_{ij}
 (\bm L_i\bm L_i^T+[\bm k_i]_\times).
\end{aligned}
\label{eq:app_soft_spin_neighbor_block}
\end{equation}
We minimize the conditional-score loss~\eqref{eq:deltaM_neural_loss} with
equal weights for all correlations and fitted lags, dividing the mean
squared residual by \((3.856\times10^{-2})^2\). No mean-mobility penalty
is imposed. \revc{The operator is estimated from \(2\times10^5\) pairs per fitted lag,
with a further \(5\times10^4\) held out. Training uses Adam with weight decay
\(10^{-7}\) and a learning rate that rises to \(3\times10^{-4}\) and decays by
cosine annealing to \(3\times10^{-6}\), over \(150\) epochs of \(64\) steps;
each step uses \(16384\) pairs at each of two randomly drawn lags. The checkpoints
saved every \(25\) epochs and the one with the smallest held-out \(E_A\), defined
below, are compared by simulating the targets with \(288\) trajectories of
\(60\) time units. The retained checkpoint, at epoch \(25\), gives the smallest
mean target-correlation error.}

\revc{The integration-by-parts test~\eqref{eq:app_conditional_ibp}, with \(h=1\)
and \(h=m_{i,c}\) and with \(\psi\) the coordinates of a site and its neighbor,
\(\varrho_i\), and \(m_{i,z}^2\), gives \(R=0.082\) and \(0.053\) over the fitted
lags, consistent with sampling noise.} To assess the conditional-score error
\revc{on the calibration constraints}, we evaluate the relative discrepancy
\begin{equation}
 E_A(\bm M)=
 \frac{\left\|\widehat{\mathcal A}
 [\bm M-\bm\Phi]-\widehat{\bm y}\right\|_2}
 {\|\widehat{\bm y}\|_2},
\end{equation}
where the constraints are stacked over the indicated observable families and
fitted lags. The results are
\begin{equation}
\begin{array}{c|cc}
 &\bm M_{\mathrm{true}}&\bm M_\theta\\ \hline
 \text{all families}&\revc{0.039}&\revc{0.048}\\
 b_1(\varrho_i)\bm m_i\text{ family}&\revc{0.16}&\revc{0.37}
\end{array}
\end{equation}
For comparison, we evaluate the calibration equations using the true
mobility and the learned stationary score, through the stationary-score
identity. The relative discrepancy over all eleven observable families and
fitted lags is \revc{\(1.1\%\)}, compared with \revc{\(3.9\%\)} when the learned
conditional score is used. Even with the true mobility, using the learned
conditional score gives correlation derivatives that differ from those
estimated directly from the data. These discrepancies are especially large
for observables involving fluctuations in spin length
(Fig.~\ref{fig:soft_spin_residuals}). Errors in these constraints can
therefore bias the recovered longitudinal diffusion.

\paragraph*{Field errors.}
Let \(\widehat{\bm D}=(\widehat{\bm M}+\widehat{\bm M}^T)/2\) and
\(\widehat{\bm R}=(\widehat{\bm M}-\widehat{\bm M}^T)/2\) denote the
symmetric and antisymmetric parts of the fitted onsite mobility, and let
\(\bm n=\bm m/\|\bm m\|\). We compare the diffusion along the spin, the
average diffusion in the two perpendicular directions, and the component of
the antisymmetric mobility proportional to \([\bm m]_\times\):
\begin{equation}
\begin{aligned}
 \widehat\lambda_\parallel&=\bm n^T\widehat{\bm D}\bm n,\\
 \widehat\lambda_\perp&=\tfrac12\left(\operatorname{tr}\widehat{\bm D}
                              -\widehat\lambda_\parallel\right),\\
 \widehat c_\times&=
 \frac{\langle\widehat{\bm R},[\bm m]_\times\rangle_F}
      {\|[\bm m]_\times\|_F^2}.
\end{aligned}
\end{equation}
For the true mobility, these quantities are the longitudinal and transverse
diffusion eigenvalues and the precession coefficient.

The maps are evaluated at \(\bm m=(m_\perp,0,m_z)\) on a
\(121\times91\) grid covering \(-1.25\leq m_z\leq1.25\) and
\(0\leq m_\perp\leq1.25\). When evaluating the neural mobility, all
neighboring spins are assigned the same value \(\bm m\). To weight the
errors, each sampled spin state is represented by
\((m_z,m_\perp)\), with \(m_\perp=(m_x^2+m_y^2)^{1/2}\), and assigned to
the nearest grid point. The weight \(w\) at each grid point is the fraction
of samples assigned to it, pooling all sites and sampled times from all
trajectories. The point \(\bm m=0\), where \(\bm n\) is undefined, has
negligible weight. Errors in more frequently observed spin states therefore
contribute more to
\begin{equation}
 e_f=\left[
 \frac{\sum_{\mathrm{grid}}w(\widehat f-f_{\mathrm{true}})^2}
      {\sum_{\mathrm{grid}}w f_{\mathrm{true}}^2}\right]^{1/2},
 \qquad f\in\{\lambda_\perp,\lambda_\parallel,c_\times\}.
\label{eq:soft_spin_field_error}
\end{equation}
Thus \(e_f\) is the weighted root-mean-square field error divided by the
weighted root-mean-square magnitude of the true field. For precession,
\(c_{\times,\mathrm{true}}=-0.8\) is constant, so the denominator is
\(0.8\). \revc{The drift error of Fig.~\ref{fig:spin_km} is the root-mean-square
difference between the drift of a model and the exact drift over the sampled
states, divided by the root-mean-square exact drift. For the true mobility with
the network score, it is \(0.13\%\).}

\end{revbtext}
\end{revtext}

\begin{figure*}[!htb]
 \centering
 \includegraphics[width=\textwidth]{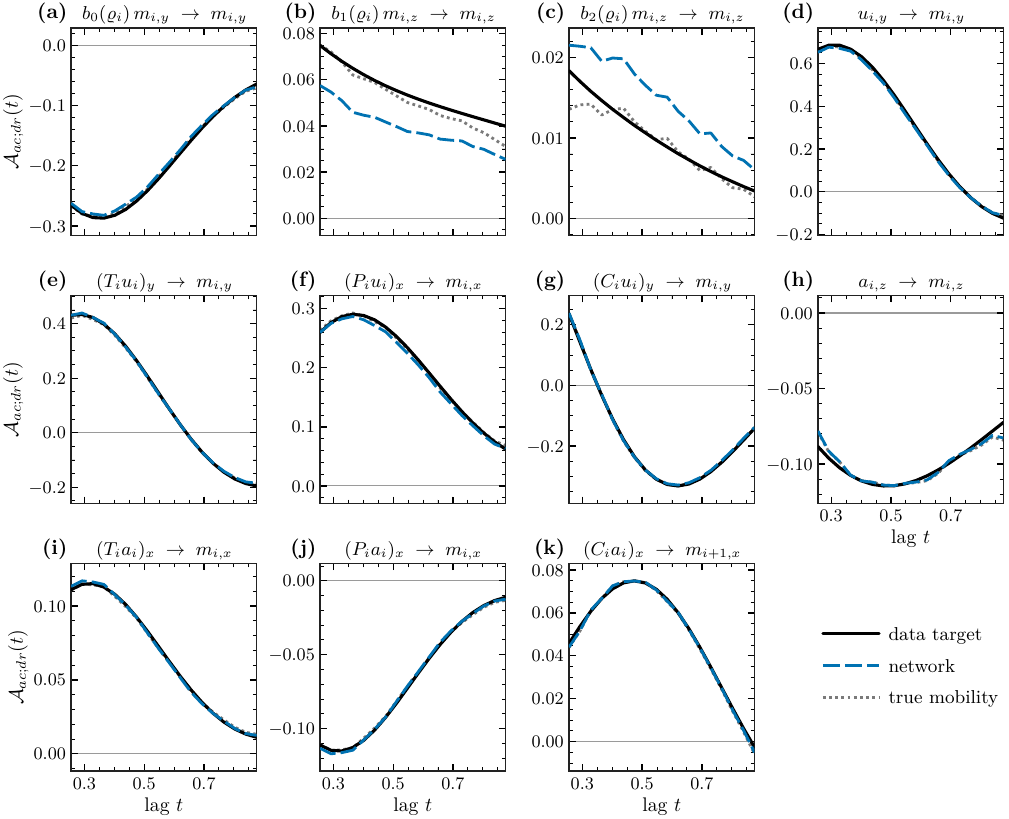}
 \caption{\revb{Conditional-score residual-operator check for the soft-spin
 chain. Each panel shows the channel of largest norm in one of the eleven
 families of Eq.~\eqref{eq:soft_spin_eleven_family_library}, over the fitted
 lags. Black: data target. Blue dashed: neural mobility. Gray dotted: true
 mobility evaluated with the same learned conditional-score operator.
 The discrepancy between gray and black measures the operator's error for
 the known mobility.}}
 \label{fig:soft_spin_residuals}
\end{figure*}

\clearpage
\setlength{\bibsep}{3pt plus 1pt}
\bibliographystyle{unsrtnat}
\bibliography{references}
\end{document}